\documentclass[10pt]{article}
\usepackage{amsmath}
\usepackage{amssymb}
\usepackage{times}
\usepackage{graphicx}
\usepackage{color}
\usepackage{multirow}
\usepackage{url}
\usepackage{hyperref}
\usepackage{listings}
\usepackage{rotating}
\usepackage{bbm}
\usepackage{latexsym}

\newcommand{\captionfonts}{\normalsize}

\makeatletter  
\long\def\@makecaption#1#2{%
  \vskip\abovecaptionskip
  \sbox\@tempboxa{{\captionfonts #1: #2}}%
  \ifdim \wd\@tempboxa >\hsize
    {\captionfonts #1: #2\par}
  \else
    \hbox to\hsize{\hfil\box\@tempboxa\hfil}%
  \fi
  \vskip\belowcaptionskip}
\makeatother   

\definecolor{codegreen}{rgb}{0,0.6,0}
\definecolor{codegray}{rgb}{0.5,0.5,0.5}
\definecolor{codepurple}{rgb}{0.58,0,0.82}
\definecolor{backcolour}{rgb}{0.95,0.95,0.92}

\lstdefinestyle{mystyle}{
    backgroundcolor=\color{backcolour},   
    commentstyle=\color{codegreen},
    keywordstyle=\color{magenta},
    numberstyle=\tiny\color{codegray},
    stringstyle=\color{codepurple},
    basicstyle=\ttfamily\footnotesize,
    breakatwhitespace=false,         
    breaklines=true,                 
    captionpos=b,                    
    keepspaces=true,                 
    numbers=left,                    
    numbersep=5pt,                  
    showspaces=false,                
    showstringspaces=false,
    showtabs=false,                  
    tabsize=2
}

\newcommand{\eps}{{\epsilon}}
\newcommand{\ld}{\mathrm{ld}}
\newcommand{\pr}{\mathrm{pr}}

\newcommand{\setR}{{\mathbb{R}}}

\newcommand{\bBox}{\hfill \mbox{$\Box$}}

\begin{document}
\hspace{13.9cm}1

\ \vspace{20mm}\\

{\LARGE Neural auto-association with optimal Bayesian learning}

\ \\
{\bf \large Andreas Knoblauch$^{\displaystyle 1}$}\\
{$^{\displaystyle 1}$Albstadt-Sigmaringen University, KEIM Institute, Poststrasse 6, 72458 Albstadt-Ebingen, Germany, knoblauch@hs-albsig.de}\\
%


\thispagestyle{empty}
\markboth{}{NC instructions}
\ \vspace{-0mm}\\
%
\begin{center} {\bf Abstract} \end{center}
Neural associative memories are single layer perceptrons with fast 
synaptic learning typically storing discrete associations between pairs 
of neural activity patterns. Previous works have analyzed the optimal networks
under naive Bayes assumptions of independent pattern components and heteroassociation,
where the task is to learn associations from input to output patterns. Here I study
the optimal Bayesian associative network for auto-association where input and output
layers are identical. In particular, I compare performance to different variants
of approximate Bayesian learning rules, like the BCPNN (Bayesian Confidence Propagation Neural Network),
and try to explain why sometimes the suboptimal learning rules achieve higher storage capacity
than the (theoretically) optimal model. It turns out that performance can depend on subtle
dependencies of input components violating the ``naive Bayes'' assumptions. This includes
patterns with constant number of active units, iterative retrieval where patterns are repeatedly propagated through
recurrent networks, and winners-take-all activation of the most probable units. Performance of all learning rules can improve significantly if
they include a novel adaptive mechanism to estimate noise in iterative retrieval steps (ANE).
The overall maximum storage capacity is achieved again by the Bayesian learning rule with ANE. 
\tableofcontents

\newpage

\section{Introduction}  \label{sec:introduction}

Neural associative memory is an alternative computing architecture in which, unlike to the classical von Neumann machine \cite{Burks/Goldstine/vonNeumann:1946,Backus:1978},
computation and data storage is not separated \cite{Steinbuch:1961,Willshaw/Buneman/Longuet-Higgins:1969,Palm:1980,Palm:1982,Hopfield:1982,Palm/Palm:1991,Knoblauch/Palm/Sommer:NeurComp2010,Knoblauch:NeurComp2016,Krotov/Hopfield:2016}. Given a training data set of $M$ associations $\{(u^{\mu}\rightarrow v^{\mu}): \mu=1,...,M\}$
between inputs $u^{\mu}$ and outputs $v^{\mu}$, the general ``hetero-associative'' task is to find, for a new input query pattern $\tilde{u}$, the most similar
$u^\mu$ from the training data and return the associated output pattern $\hat{v}=v^\mu$. This is similar to nearest-neighbor algorithms \cite{Fix/Hodges:KNN:1951,Cover/Hart:KNN:1967,Knoblauch:2005_b,Gripon/Lowe/Vermet:2018} and 
locality-sensitive hashing \cite{Indyk/Motwani/Raghavan/Vempala:1997,Indyk/Motwani:1998,Motwani/Naor/Panigrahy:2006}, but neural implementations turned out to be particularly efficient for some applications
and play an important role in neuroscience as models of neural computation 
for various brain structures, for example neocortex, hippocampus, cerebellum, mushroom body 
\cite{Hebb:1949,Braitenberg:1978,Palm:1982,Fransen/Lansner:1998,Pulvermuller:2003_a,Johansson/Lansner:2007,Lansner:2009,Marr:1971,Gardner-Medwin:1976,Rolls:1996,Bogacz/Brown/Giraud-Carrier:2001,Marr:1969,Albus:1971,Kanerva:1988,Laurent:2002,Knoblauch/Markert/Palm:2005_b,Fay/Kaufmann/Knoblauch/Markert/Palm:2005,Knoblauch_IAMCOIL:HRI2008,Palm/Knoblauch/Hauser/Schuz:BiolCyb2014}.
A special case is ``auto-association'' where inputs and outputs are identical, $u^{\mu}=v^{\mu}$, thus associating
each input $u^{\mu}$ with itself \cite{Willshaw/Buneman/Longuet-Higgins:1969,Palm:1980,Hopfield:1982}.
Corresponding applications involve, for example, pattern completion (where $\tilde{u}$ is a noisy version of $u^{\mu}$), clustering, and modeling the
recurrent architecture of brain networks implementing Hebbian cell assemblies \cite{Hebb:1949,Palm:1982,Palm/Knoblauch/Hauser/Schuz:BiolCyb2014,Lansner/Ravichandran/Herman:2023}. 

In its simplest forms, neural associative memories are single layer perceptrons \cite{Rosenblatt:1958} that employ fast one-shot synaptic learning. This favors most plausibly local learning rules 
for synaptic weights $w_{ij}$ that depend only on presynaptic and postsynaptic activities $u^\mu_i$ and $v^\mu_j$ ``visible''
to the synapse \cite{Steinbuch:1961,Willshaw/Buneman/Longuet-Higgins:1969,Palm:1980,Hopfield:1982,Diederich/Opper:1987,Lansner/Ekeberg:1987,Lansner/Ekeberg:1989,Willshaw/Dayan:1990,Palm:1991,Palm:1992,Knoblauch:Elsevier2017}.
In previous works I have investigated the optimal local learning rule under a ``naive Bayes'' assumption that input pattern components $u^\mu_i$ are statistically
independent \cite{Knoblauch_BayesAsso:HRI2009,Knoblauch_BayesAssoIJCNN:2010,Knoblauch_LansnerAsso:HRI2010,Knoblauch:NeurComp2011}.
This Bayesian learning rule has been shown to be optimal, achieving minimal output noise (as measured in terms of expected Hamming distance between the output $\hat{v}$ and the original output $v^{\mu}$)
and maximal storage capacity (as evaluated in terms of the maximal pattern number that can be stored at a given output noise level)
in a hetero-associative one-step-retrieval setting, where the output $\hat{v}$ is computed from a single sweep propagating the input query pattern $\tilde{u}$ through the network.

In this technical report, I study the auto-associative version of the optimal Bayesian learning rule. The motivation for this work was triggered by a personal communication from Anders Lansner \cite{Lansner:2024:perscomm}
who observed that the BCPNN learning rule (BCPNN=Bayesian Confidence Propagation Neural Network \cite{Lansner/Ekeberg:1987,Lansner/Ekeberg:1989,Lansner/Holst:1996,Johansson/Lansner:2007,Lansner/Ravichandran/Herman:2023})
performed better than the ``optimal'' Bayesian rule in some auto-associative scenarios.
This raised some questions about the correctness of the implementation of the optimal Bayesian learning rule (as he used my implementation from \cite{Knoblauch:NeurComp2011}) and the applicability of the hetero-associative theory
of \cite{Knoblauch:NeurComp2011} to auto-associative networks.
Therefore, sections~\ref{sec:learning} and ~\ref{sec:retrieval_onestep} reformulate and scrutinize the theory of \cite{Knoblauch:NeurComp2011}
for the special case of auto-association. It gives also several approximations of the optimal Bayesian that, including the BCPNN learning rule, and discuss conditions of convergence. It is concluded that,
if properly implemented, the theory of \cite{Knoblauch:NeurComp2011} should hold as well for auto-associative networks. Then, in order to find an alternative explanation for the anomaly that a suboptimal learning rule
like BCPNN can perform better than the ``optimal'' Bayesian rule,
section~\ref{sec:experiments} shows results from numerical experiments comparing the optimal Bayesian rule to BCPNN using the implementation of \cite{Knoblauch:NeurComp2011}.
The experiments reveal that the anomaly occurs only in an iterative retrieval scenario \cite{Schwenker/Sommer/Palm:1996}
with a winners-take-all activity selection, if using memory patterns that do not strictly fulfill the independence assumptions of the optimal Bayesian model (e.g., patterns that have a fixed number of active units).
Accordingly, section~\ref{sec:discussion} concludes that the hypothesis of an incorrect implementation of optimal Bayesian learning
can be rejected. Instead the results show (once more) that theories on the storage capacity of neural associative networks strongly depend on independence assumptions that are seldom fulfilled in real-world
applications \cite{Knoblauch_IAMCOIL:HRI2008}, and even subtle deviations from these assumptions can invalidate theoretical optimality predictions. 

\newpage

\section{Memory storage in neural and synaptic counter variables for auto-association}\label{sec:learning}

The task is to store $M$ auto-associations of activity patterns ${\bf u}^\mu$,  
where $\mu=1\ldots M$. Here the ${\bf u}^\mu$ are binary vectors of size $n$. 
Associations are stored in first order (neural) and second order (synaptic) counter variables.
In particular, each neuron $i$ can memorize its {\it unit usage} 
\begin{eqnarray}
  M_1 (i) & := & \#\{\mu: u^{\mu}_i = 1\}             \label{eq:M1} \\
  M_0 (i) & := & \#\{\mu: u^{\mu}_i = 0\} = M-M_1(i)  \label{eq:M0}
\end{eqnarray}
Similarly, each synapse $ij$ can memorize its {\it synapse usage}   
\begin{eqnarray}
  M_{11} (ij) & := & \#\{\mu: u^{\mu}_i = 1, u^{\mu}_j = 1\} = M_{11}(ji)                      \label{eq:M11} \\
  M_{01} (ij) & := & \#\{\mu: u^{\mu}_i = 0, u^{\mu}_j = 1\} = M_{10}(ji) = M_1(j)-M_{11}(i,j)  \label{eq:M01} \\
  M_{00} (ij) & := & \#\{\mu: u^{\mu}_i = 0, u^{\mu}_j = 0\} = M_{00}(ji) = M_0(i)-M_{01}(i,j)  \label{eq:M00} \\
  M_{10} (ij) & := & \#\{\mu: u^{\mu}_i = 1, u^{\mu}_j = 0\} = M_{01}(ji) = M_0(j)-M_{00}(i,j)  \label{eq:M10} 
\end{eqnarray}
where $i,j=1,\ldots,n$. For autapses with $i=j$ it holds
\begin{eqnarray}
  M_{11} (ii) = M_{1}(i)\ , \quad M_{00} (ii) = M_{0}(i)\ , \quad M_{01} (ii)=M_{10} (ii)=0\ .  \label{eq:Muu_autapses}
\end{eqnarray}
Note that it is sufficient to memorize $M$ and $M_{11}(ij)$ for $i\ge j$.
Thus, an implementation on a digital computer requires about $(\frac{n(n+1)}{2}+1)\ld M$ memory bits. 
The following analyses consider optimal Bayesian retrieval assuming that each neuron $j=1,\ldots,n$ has access to the 
variables in the set
\begin{eqnarray}
  \mathfrak{M}(j) & := & \{ M, M_1(j), M_{11}(ij): i=1,\ldots,n\}\ .   \label{eq:fracM_j}
\end{eqnarray}

\section{Neural formulation of optimal Bayesian one-step retrieval for auto-association}\label{sec:retrieval_onestep}

Given a query pattern $\tilde{u}$ and the counter variables of section~\ref{sec:learning}, 
the memory task is to find the ``most similar'' address
pattern ${\bf u}^\mu$ and return a corresponding reconstruction ${\bf\hat{u}}$.  
In general, the query $\tilde{u}$ is a noisy version of a stored pattern
$u^\mu$, assuming component transition probabilities given the activity of the original pattern, $u^\mu_j=\frak{a}\in\{0,1\}$,
\begin{eqnarray}
  p_{01|\frak{a}}(ij) &:=& \pr[\tilde{u}_i=1|u^\mu_i=0,u^\mu_j=\frak{a}] \ ,\label{eq:p01a}\\
  p_{10|\frak{a}}(ij) &:=& \pr[\tilde{u}_i=0|u^\mu_i=1,u^\mu_j=\frak{a}] \label{eq:p10a} \ ,
\end{eqnarray}
where for autapses with $i=j$ only $p_{01|0}(ii)$ and $p_{10|1}(ii)$ are relevant. Although irrelevant, we may define
\begin{eqnarray}
  p_{01|1}(ii):=0\ , \quad p_{10|0}(ii):=0  \label{eq:p01a_p10a_irrelevant}
\end{eqnarray}
for convenience (see below). Now the neurons $j$ have to decide independently of each other 
whether to be activated or to remain silent. Given the query $\tilde{u}$, 
the optimal maximum-likelihood decision based on the odds ratio $\frak{r}_j$ is
\begin{eqnarray}
  \hat{u}_j & = & 
  \left\{ \begin{array}{c@{\:,\quad}l}
    1 &  \frak{r}_j := \frac{\pr[u^\mu_j=1|{\tilde{u}},\frak{M}(j)]}{\pr[u^\mu_j=0|{\tilde{u}},\frak{M}(j)]} \ge 1   \\
    0 &  {\rm otherwise}     \\
  \end{array} \right. \ , \label{eq:retrieval}
\end{eqnarray}
minimizing the expected Hamming distance $d_H(u^\mu,\hat{u}):=\sum_{j=1}^n|u^\mu_j-\hat{u}_j|$ between original and reconstructed pattern.  
If the query pattern components are conditional independent given the activity of neuron $j$, e.g., 
assuming independently generated pattern components, we have for $\frak{a}\in\{0,1\}$
\begin{eqnarray}
  \pr[\tilde{u}|u^\mu_j=\frak{a},\frak{M}(j)] & = & \prod_{i=1}^n \pr[\tilde{u}_i|u^\mu_j=\frak{a},\frak{M}(j)] \nonumber \\
            & = & \prod_{i=1}^n 
  \frac{M_{\tilde{u}_i\frak{a}}(ij)(1-p_{\tilde{u}_i(1-\tilde{u}_i)|\frak{a}}(ij)) + M_{(1-\tilde{u}_i)\frak{a}}(ij)p_{(1-\tilde{u}_i)\tilde{u}_i|\frak{a}}(ij)}
       {M_\frak{a}{(j)}} \ , \label{eq:p_u_tilde_given_a_naiveBayes}
\end{eqnarray}
where using (\ref{eq:p01a_p10a_irrelevant}) turns out convenient (avoiding distinction of cases) due to (\ref{eq:Muu_autapses}). 
With the Bayes formula $\pr[u^\mu_j=\frak{a}|{\tilde{u}},\frak{M}(j)]=\pr[{\tilde{u}}|u^\mu_j=\frak{a},\frak{M}(j)]\pr[u^\mu_j=\frak{a}|\frak{M}(j)]/\pr[{\tilde{u}}|\frak{M}(j)]$ 
and $\pr[u^\mu_j=\frak{a}|\frak{M}(j)]=M_{\frak{a}}(j)/M$, the odds ratio is
\begin{eqnarray}
 \frak{r}_j  =  \left(\frac{M_0(j)}{M_1(j)}\right)^{n-1} \prod_{i=1}^n
        \frac{M_{\tilde{u}_i1}(ij)(1-p_{\tilde{u}_i(1-\tilde{u}_i)|1}(ij)) + M_{(1-\tilde{u}_i)1}(ij)p_{(1-\tilde{u}_i)\tilde{u}_i|1}(ij)}
        {M_{\tilde{u}_i0}(ij)(1-p_{\tilde{u}_i(1-\tilde{u}_i)|0}(ij)) + M_{(1-\tilde{u}_i)0}(ij)p_{(1-\tilde{u}_i)\tilde{u}_i|0}(ij)}\ .  \label{eq:fpp} 
\end{eqnarray}
For a more plausible neural formulation, we can take logarithms of the probabilities and obtain
synaptic weights $w_{ij}$ and dendritic potentials $x_j:=\log\frak{r}_j$, 
\begin{eqnarray}
  w_{ij} & = & \log\frac{(M_{11}(1-p_{10|1})+M_{01}p_{01|1})(M_{00}(1-p_{01|0})+M_{10}p_{10|0})}{(M_{10}(1-p_{10|0})+M_{00}p_{01|0})(M_{01}(1-p_{01|1})+M_{11}p_{10|1})} \label{eq:wij_Bayesian} \\
  x_j    & = & (n-1)\log\frac{M_0}{M_1} + \sum_{i=1}^n \log\frac{M_{01}(1-p_{01|1})+M_{11}p_{10|1}}{M_{00}(1-p_{01|0})+M_{10}p_{10|0}} 
                 + \sum_{i=1}^n w_{ij}\tilde{u}_i \label{eq:xj_Bayesian}
\end{eqnarray}
such that $\pr[u^\mu_j=1|\tilde{u},\frak{M}(j)]=1/(1+e^{-x_j})$ writes as a sigmoid function of $x_j$,  
and neuron $j$ fires, $\hat{u}_j=1$, iff the dendritic potential is non-negative, $x_j\ge 0$. 
Note that we have skipped indices of $M_0(j)$, $M_1(j)$, $p_{01}(ij)$, $p_{10}(ij)$, 
$M_{00}(ij)$, $M_{01}(ij)$, $M_{10}(ij)$, $M_{11}(ij)$ for the sake of readability.
Note the following points: 
\begin{itemize}
\item Evaluating eq.~\ref{eq:xj_Bayesian} is {\bf computationally much cheaper} than using eq.~\ref{eq:fpp},
      in particular for sparse queries having only a small number of active components with $\tilde{u}_i=1$. 
\item Optimal Bayesian learning for auto-association is a {\bf non-linear learning rule}, that is, due to the logarithms,
      the synaptic weights $w_{ij}$ and biases in $x_j$ are non-linear functions of the synaptic counters $M_{uv}(ij)$.
\item Note that, for the sake of convenience, equations (\ref{eq:p_u_tilde_given_a_naiveBayes}-\ref{eq:xj_Bayesian}) use
      the {\bf ill-defined error probabilities} $p_{01|1}(ii)$ and $p_{10|0}(ii)$
      of (\ref{eq:p01a_p10a_irrelevant}) for {\bf autapses} with $i=j$. Their values are actually irrelevant (and can be chosen arbitrarily),
      because they always get multiplied with zero-valued synaptic counters $M_{01}(ii)=M_{10}(ii)=0$ according to (\ref{eq:Muu_autapses}).
      Thus, the definitions (\ref{eq:p01a_p10a_irrelevant}) could be replaced by {\bf arbitrary values} $p_{01|1}(ii),p_{10|0}(ii)\in\setR$.
\item In general, the {\bf recurrent weight matrix} is {\bf asymmetric}, that is $w_{ij}\neq w_{ji}$. A {\bf symmetric weight matrix} with $w_{ij}=w_{ji}$
      is obtained only only if assuming {\bf zero error probabilities} $p_{01|\frak{a}}=p_{10|\frak{a}}=0$, due to the symmetry relations
      in (\ref{eq:M11}-\ref{eq:M10}).
\item To satisfy {\bf Dale's law} that all synapses of the neurons are (either) excitatory (or inhibitory), we may add
      a sufficiently large constant $c:=-\min_{ij}w_{ij}$ to each weight. 
      Then all synapses have non-negative weights $w_{ij}':=w_{ij}+c\ge 0$ and the dendritic potentials remain unchanged if we replace the last sum in
      eq.~\ref{eq:xj_Bayesian} by
      \begin{eqnarray}
         \sum_{i=0}^m w_{ij}\tilde{u}_i = \sum_{i=0}^m w_{ij}'\tilde{u}_i -c\sum_{i=0}^m \tilde{u}_i  \ .\label{eq:xj_Bayesian_Dale}
      \end{eqnarray}
      Here the negative sum could be realized, for example, by feed-forward inhibition 
      with a strength proportional to the query pattern activity, e.g., as suggested by 
      \cite{Knoblauch/Palm:2001_a,Knoblauch:2005_b}.
\item The {\bf error probabilities} eqs.~\ref{eq:p01a},\ref{eq:p10a} can be estimated by maintaining counter variables
      similar as in section~\ref{sec:learning}. 
      For example, if the $\mu$th memory $u^\mu$ has been queried by ${\tilde M}^\mu$ address queries $\tilde{u}^{(\mu,\mu')}$ 
      (where $\mu'=1,2,\ldots,\tilde{M}^\mu$) then we could estimate for $\frak{a},\frak{b},\frak{c}\in\{0,1\}$
      \begin{eqnarray}
      p_{\frak{b}\frak{c}|\frak{a}}(ij) = \frac {\#\{(\mu,\mu'): u^{\mu}_i = \frak{b}, \tilde{u}^{(\mu,\mu')}=\frak{c}, u^{\mu}_j = \frak{a}\}} 
                                            {\#\{(\mu,\mu'): u^{\mu}_i = \frak{b}, u^{\mu}_j = \frak{a}, 1\le \mu'\le \tilde{M}^\mu\}} 
      \end{eqnarray}
      which requires four counter variables per synapse in addition to $M_{11}$. To {\bf reduce storage costs} one may assume
      \begin{eqnarray}
       p_{\frak{b}\frak{c}|\frak{a}}(ij) = p_{\frak{b}\frak{c}}(i) = \frac {\#\{(\mu,\mu'): u^{\mu}_i = \frak{b}, \tilde{u}^{(\mu,\mu')}=\frak{c}\}} 
                                            {\#\{(\mu,\mu'): u^{\mu}_i = \frak{b}, 1\le \mu'\le \tilde{M}^\mu\}} 
      \end{eqnarray}
      independent of $j$ as do the following analyses for the sake of simplicity, although this assumption may 
      reduce the number of discovered rules (corresponding to infinite $w_{ij}$) 
      describing deterministic relationships between $u_i$ and $u_j$. The {\bf simplest method} would be to estimate
      \begin{eqnarray}
       p_{\frak{b}\frak{c}|\frak{a}}(ij) = p_{\frak{b}\frak{c}} = \frac {\#\{(\mu,\mu',i): u^{\mu}_i = \frak{b}, \tilde{u}^{(\mu,\mu')}=\frak{c}\}} 
                                            {\#\{(\mu,\mu',i): u^{\mu}_i = \frak{b}, 1\le \mu'\le \tilde{M}^\mu\}}   \label{eq:error_prob_indep_of_a}
      \end{eqnarray}
      independent of $j$ and $i$, such that we have to store only two error probabilities $p_{01}$ and $p_{10}$.
      For mean activity $k:=E(\sum_{i=1}^nu_i^{\mu})$ we may use instead the {\bf input noise parameters} $\lambda$ and $\kappa$ with
      \begin{eqnarray}
        \lambda&:=1-p_{10}  \quad\quad\left(\ \mbox{or}\quad p_{10}=1-\lambda\  \right)                        \label{eq:lambda}\\
        \kappa&:=\frac{(n-k)p_{01}}{k} \quad\quad\left(\ \mbox{or}\quad p_{01}=\frac{\kappa k}{n-k}\  \right)  \label{eq:kappa}
      \end{eqnarray}
      which express the fractions of the mean numbers $\lambda k$ and $\kappa k$ of ``correct'' and ``false'' one-entries
      in a query pattern $\tilde{u}$, respectively.
\item In the limit of {\bf sparse patterns} the fraction of active units
      \begin{eqnarray}
         p(i) := \pr[u_i^{\mu}=1] \approx \frac{1}{M}\sum_{\mu=1}{M}u_i^{\mu} \rightarrow 0     \label{eq:sparselimit_p_0}
      \end{eqnarray}
      vanishes for large networks with $n\rightarrow\infty$. Correspondingly,
      $M_{11}\ll M_{10},M_{01} \ll M_{00}$, such that (\ref{eq:wij_Bayesian},\ref{eq:xj_Bayesian}) with (\ref{eq:error_prob_indep_of_a}) become
      \footnote{
        The  bias in (\ref{eq:xj_Bayesian_sparse}) follows from  
        $(n-1)\log\frac{M_0}{M_1} + \sum_{i=1}^n \log\frac{M_{01}(1-p_{01|1})+M_{11}p_{10|1}}{M_{00}(1-p_{01|0})+M_{10}p_{10|0}}$
        $\simeq (n-1)\log\frac{M_0}{M_1} + \sum_{i=1}^n \log\frac{M_{01}}{M_{00}} \simeq (n-1)\log\frac{M_0}{M_1} + \sum_{i=1}^n \log\frac{M_{01}+M_{11}}{M_{00}+M_{10}}$
        $=(n-1)\log\frac{M_0}{M_1} + n\log\frac{M_{1}}{M_{0}}=\log\frac{M_{1}}{M_{0}}$. \bBox
      }
      \begin{eqnarray}
         w_{ij} & \simeq & \log\frac{(M_{11}(1-p_{10})+M_{01}p_{01})M_{00}}{(M_{10}(1-p_{10})+M_{00}p_{01})M_{01}} \label{eq:wij_Bayesian_sparse} \\
         x_j    & \simeq & \log\frac{M_1}{M_0} + \sum_{i=1}^n w_{ij}\tilde{u}_i \ .\label{eq:xj_Bayesian_sparse}
      \end{eqnarray}
\item In the limit of {\bf sparse patterns} with sufficiently {\bf low add-noise} $p_{01}\rightarrow 0$ such that $M_{01}p_{01}\ll M_{11}(1-p_{10})$ and $M_{00}p_{01}\ll M_{10}(1-p_{10})$,
      (\ref{eq:wij_Bayesian_sparse}) becomes simply
      \begin{eqnarray}
         w_{ij} & \simeq & \log\frac{M_{11}M_{00}}{M_{10}M_{01}}\ . \label{eq:wij_Bayesian_sparse_lowaddnoise}
      \end{eqnarray}
      which asymptotically equals the {\bf original BCPNN learning rule} \cite{Lansner/Ekeberg:1987,Lansner/Ekeberg:1989}
      \footnote{
        Assuming sparse patterns with low add-noise (\ref{eq:wij_Bayesian_sparse_lowaddnoise}) becomes $w_{ij} \simeq \log\frac{(M_{11}M_{00}}{M_{10}M_{01}}$
        $\simeq \log\frac{(M_{11}(M_{00}+M_{01}+M_{10}+M_{11})}{(M_{10}+M_{11})(M_{01}+M_{11})}=\log\frac{M_{11}M}{M_{1}(i)M_1(j)}=w_{ij}^{\mathrm{BCPNN}}$. \bBox
      }
      \begin{eqnarray}
          w_{ij}^{\mathrm{BCPNN}} & = & \log\frac{M_{11}(ij)M}{M_1(j)M_1(i)}  \label{eq:wij_Bayesian_BCPNN_weights} \\
                 x_j^{\mathrm{BCPNN}}    & = & \log\frac{M_1(j)}{M}   + \sum_{i=1}^n w_{ij}\tilde{u}_i \ .\label{eq:xj_Bayesian_BCPNN_potentials}
      \end{eqnarray}
      This means that {\bf BCPNN is optimal for sparse patterns and low add-noise}. For example, for {\bf i.i.d. pattern components} with $p:=p(i)\rightarrow 0$, this requires
      \footnote{
        For i.i.d. pattern components $u_i^{\mu}=1$ with $p:=\pr[u_i^{\mu}=1]$ the spike counters are binomials, where expectations $\pm$ standard deviations are
        $M_{11}\simeq Mp^2\pm \sqrt{M}p$, $M_{01},M_{10}\simeq Mp\pm\sqrt{Mp}$, $M_{00}\simeq M\pm \sqrt{M}$. Thus $p\rightarrow 0$ implies $M_{11}\ll M_{10},M_{01} \ll M_{00}$ as required for the {\bf sparse pattern limit}
        (\ref{eq:wij_Bayesian_sparse},\ref{eq:xj_Bayesian_sparse}), but the {\bf low add-noise limit} requires additionally $M_{01}p_{01}\ll M_{11}(1-p_{10})$ and $M_{00}p_{01}\ll M_{10}(1-p_{10})$,
        or $Mpp_{01}\ll Mp^2(1-p_{10})$ and $Mp_{01}\ll Mp(1-p_{10})$, or $p_{01}\ll p(1-p_{10})$. The second inequality in (\ref{eq:condition_BCPNN_optimal}) follows with (\ref{eq:lambda},\ref{eq:kappa}). \bBox
      }
      \begin{eqnarray}
         p_{01}\ll p(1-p_{10})\le p \rightarrow 0  \quad\quad\mbox{or}\quad\quad \kappa\ll\lambda\ , \label{eq:condition_BCPNN_optimal}
      \end{eqnarray}
      for example, $\kappa\rightarrow 0$ for nonvanishing $\lambda\not\rightarrow 0$. 
\end{itemize}

\section{Numerical experiments}\label{sec:experiments}

\subsection{Methods}\label{sec:experiments_methods}
Similar to \cite{Knoblauch:NeurComp2011}, the numerical experiments use the {\bf Bayesian learning rule} (``{\bf B}'') of (\ref{eq:wij_Bayesian},\ref{eq:xj_Bayesian}), and some {\bf generalizations of the BCPNN} learning rule:
To be more comparable with the Bayesian rule, the {\bf generalized BCPNN learning rule} (``{\bf BCPNN}'') extends (\ref{eq:wij_Bayesian_BCPNN_weights},\ref{eq:xj_Bayesian_BCPNN_potentials})
by estimations of query noise $p_{\frak{b}\frak{c}|\frak{a}}(ij)$ as in (\ref{eq:p01a}-\ref{eq:p01a_p10a_irrelevant}) (see \cite{Knoblauch:NeurComp2011}, app.~H.1),
\begin{eqnarray}
  -\Theta_j & = & \log 2 + \log\frac{M_1}{M} \\
  w_{ij} & := & \log\frac{(M_{11}(1-p_{10|1})+M_{01}p_{01|1})M}
                         {(M_{1}'(1-p_{10})+M_{0}'p_{01})M_1} \label{eq:wij_noisy_BCPNN}
\end{eqnarray}
where ``primed'' unit usages $M_1':=M_1(i), M_0':=M_0(i)$ correspond to presynaptic units $i$, ``unprimed'' unit usages $M_1:=M_1(j), M_0:=M_0(j)$
correspond to postsynaptic units $j$, and $-\Theta_j$ is the (negative) firing threshold (or bias as in (\ref{eq:xj_Bayesian},\ref{eq:xj_Bayesian_BCPNN_potentials})).
The additional term $\log 2$ comes from the fact that BCPNN computes the logarithm of the probability $\pr[u^\mu_j=1|1_{\tilde{u}}]$ of an original memory unit being 1
given the one-entries of the query pattern $\tilde{u}$,
which requires a decision threshold of $\frac{1}{2}$ (for details, see \cite{Knoblauch:NeurComp2011}, sect.~3.3 and app.~H.1). 
Here we define $p_{\frak{b}\frak{c}}(ij):=\pr[u^\mu_j=0]p_{\frak{b}\frak{c}|0}(ij)+\pr[u^\mu_j=1]p_{\frak{b}\frak{c}|1}(ij)$,
but in experiments we mostly use the simplifications (\ref{eq:error_prob_indep_of_a}-\ref{eq:kappa}).

As the BCPNN rule exploits only the one-entries of query patterns (\cite{Knoblauch:NeurComp2011}, sect.~3.3), we may extend it by exploiting also the zero-entries, leading to
the improved ``{\bf BCPNN2}'' learning rule (see \cite{Knoblauch:NeurComp2011}, app.~H.2)
\begin{eqnarray}
  -\Theta_j & = & \log 2 + (m-1)\log\frac{M}{M_1} 
                                    + \sum_{i=1}^m \log\frac{M_{01}(1-p_{01|1})+M_{11}p_{10|1}}{M_0'(1-p_{01})+M_1'p_{10}}\\
  w_{ij}   &=& \log\frac{(M_{11}(1-p_{10|1})+M_{01}p_{01|1})(M_0'(1-p_{01})+M_1'p_{10})}
                                 {(M_{01}(1-p_{01|1})+M_{11}p_{10|1})(M_1'(1-p_{10})+M_0'p_{01})}
  \label{eq:wij_Bayesian_BCPNNII_noisy}
\end{eqnarray}
Another extension is to include the odds ratio as in (\ref{eq:retrieval}), eliminating the need to compute $\pr[\tilde{u}]$, leading to the ``{\bf BCPNN3}'' learning rule (see \cite{Knoblauch:NeurComp2011}, app.~H.3)
\begin{eqnarray}
  -\Theta_j & = & \log\frac{M_1}{M_0} \label{eq:ThetajBCPNNIII}\\
  w_{ij}    & = & \log\frac{(M_{11}(1-p_{10|1})+M_{01}p_{01|1})M_0}
                           {(M_{10}(1-p_{10|0})+M_{00}p_{01|0})M_1} \ .\label{eq:wijBCPNNIII}
\end{eqnarray}
If naively implemented, all these Bayesian-type learning rules (\ref{eq:wij_Bayesian},\ref{eq:xj_Bayesian}) and (\ref{eq:wij_noisy_BCPNN}-\ref{eq:wijBCPNNIII}) are {\bf numerical problematic},
as weights and thresholds may become infinite if synaptic counter variables and noise estimates become zero.
However, here we use the method to compute {\bf exact results} by keeping the infinite contributions separate from the finite contributions (see \cite{Knoblauch:NeurComp2011}, app.~A).
An alternative ``{\bf dirty}'' method is it to replace each infinite contribution (factors in the logarithms that become zero) by a large real number $\pm z_{\infty}$.
\footnote{
  We did some experiments with this ``dirty'' method, using $z_{\infty}=1e+200$ as replacement for $\infty$. Perhaps an even better approach would be to choose
  $z_{\infty}$ just a bit larger than the largest dendritic potential that can occur from the finite contributions. The sign of $\pm z_{\infty}$ is positive if the contribution is from
  a term in the nominator of the logarithm, and negative if the contribution is from the denominator.
}
An even {\bf simpler dirty method} to avoid infinite values would be to just add a small constant to the denominators.
However, we will discuss below that this may lead to {\bf confusions} with {\bf prior assumptions about query noise}.
In fact, a {\bf safe method} to avoid infinite synapses and thresholds (corresponding to deterministic rules) is to explicitly choose reasonable {\bf non-zero} values for the prior {\bf query noise}
estimates (\ref{eq:p01a}-\ref{eq:p01a_p10a_irrelevant}). 

\subsection{Results for the basic models}\label{sec:experiments_results}
This work was triggered by reports that BCPNN can outperform the (theoretically optimal) Bayesian rule in a sparse regime with $k=\sqrt{n}$ \cite{Lansner:2024:perscomm}. 
In order to investigate the reported phenomenon, I reactivated the software used in \cite{Knoblauch:NeurComp2011} to investigate this phenomenon.
In a {\bf first series of experiments} I tried to {\bf reproduce the results} of \cite{Knoblauch:NeurComp2011} (Fig.4A, 5A) for heteroassociative one-step-retrieval
in order to verify the updated implementation (however, for non-sparse memory patterns with $k=n/2$).
The new data matched closely the previous data (not shown), showing the correctness of the updated implementation.

\begin{figure*}[ht]
  \begin{center}
  \includegraphics[width=\linewidth]{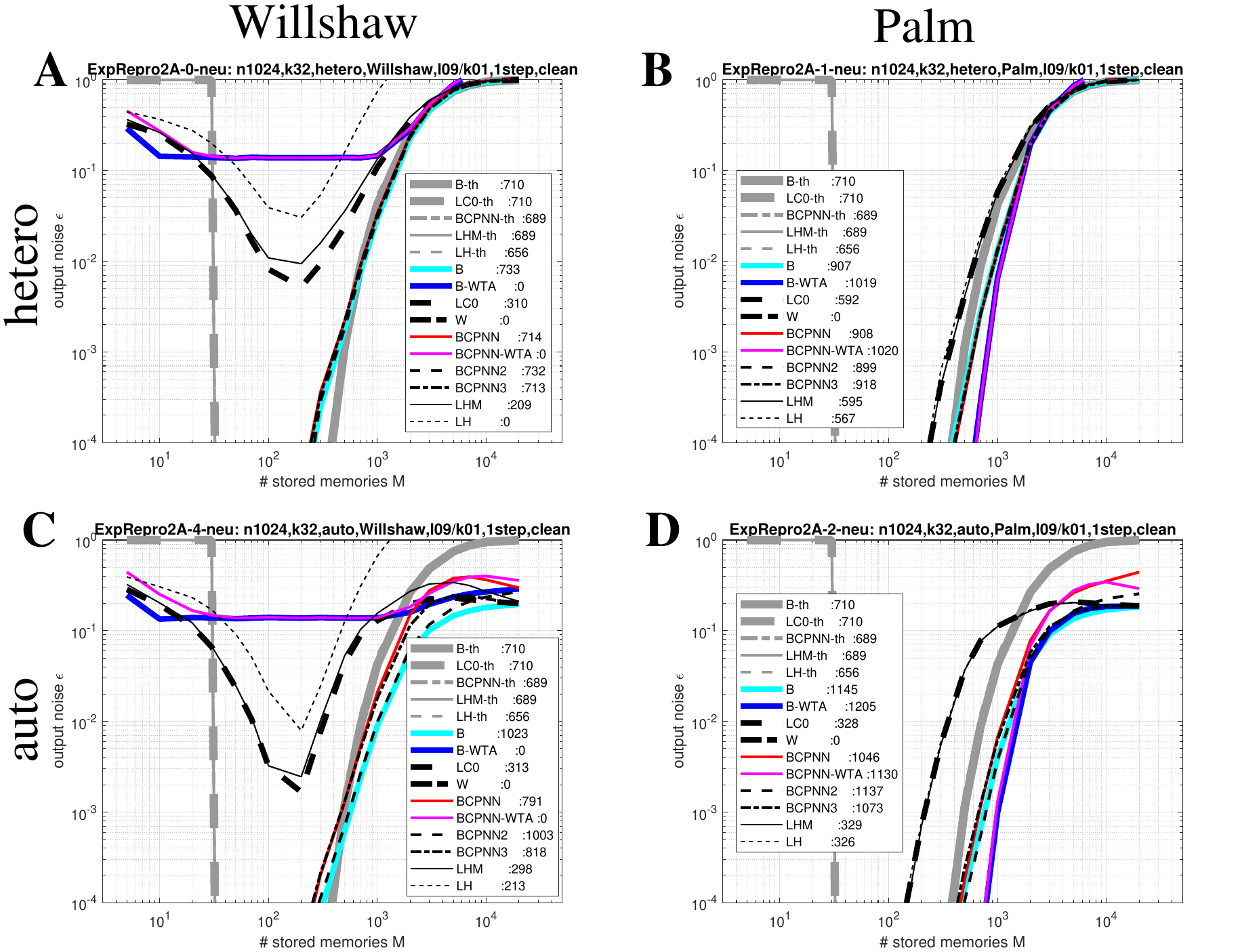}
  \end{center}
  \caption{\label{fig1:ExpRepro2A}
    Output noise $\epsilon$ as a function of stored memories for networks of size $n=1024$,
    where each pattern has $k=\sqrt{n}=32$ one-entries using {\bf one-step-retrieval} for queries with 10 percent input miss/add noise ($\lambda=0.9, \kappa=0.1$).
    {\bf Upper panels} correspond to hetero-association, {\bf lower} panels to auto-association. {\bf Left} panels correspond to Willshaw patterns
    (independent components and noise). {\bf Right} panels correspond to Palm patterns (fixed number of components and noise).
    Each panel shows results for various {\bf learning rules} (see \cite{Knoblauch:NeurComp2011} for details). In particular,
    ``B'' correspond to the Bayesian rule (\ref{eq:wij_Bayesian},\ref{eq:xj_Bayesian}),
    BCPNN to (\ref{eq:wij_noisy_BCPNN}), BCPNN2 to (\ref{eq:wij_Bayesian_BCPNNII_noisy}), BCPNN3 to (\ref{eq:wijBCPNNIII}); WTA=$k$-winners-take all retrieval; th=theory.
    The numbers in the legend correspond to (interpolated) {\bf pattern capacity} $M_{\epsilon}$ at output noise level $\eps=0.01$.
    See text for further details.
  }
\end{figure*}

A {\bf second series of experiments} tested the {\bf auto-associative models for one-step-retrieval}
and compared it to hetero-association for network size $n=1024$ and
relevant cell assembly size $k=\sqrt(n)=32$ and compared also ``{\bf Willshaw-patterns}'' (all patterns independently generated, $k$ is only mean number
of one-entries per pattern) versus ``{\bf Palm patterns}'' (all patterns have exactly $k$ one-entries; all queries have exactly
$\mathrm{round}(\lambda k)$ correct and $\mathrm{round}(\kappa k)$ false one-entries) \cite{Knoblauch:siam2008_WP,Palm:1980,Buckingham/Willshaw:1992}. Fig.~\ref{fig1:ExpRepro2A} shows results for queries with 10 percent add-noise and miss-noise (corresponding to $\lambda=0.9$ and $\kappa=0.1$), where each data point averages over 100 networks and 100 retrievals per network.
As expected, the {\bf Bayesian rule generally reaches best results}: In particularly, for Willshaw patterns, the Bayesian rule performs significantly better than the other rules.
For Palm patterns pattern capacities are significantly larger than for Willshaw patterns, and $k$-WTA (winner-takes-all) retrieval performs much better than fixed firing thresholds.
For heteroassociation, BCPNN exceeds pattern capacity of the Bayesian rule by 1 (which is not significant, however).
\footnote{\label{footnote:rounding4Palmpatterns}
  It should also be noted that due to the {\bf rounding procedure}, for {\bf Palm patterns} queries contain $\mathrm{round}(\lambda k)=\mathrm{round}(0.9\cdot 32)=\mathrm{round}(28.8)=29$ correct one-entries and
  $\mathrm{round}(0.1\cdot 32)=\mathrm{round}(3.2)=3$ false one-entries. This means that the actual query noise is actually {\bf a bit smaller than 10 percent add/miss noise}, and the
  error estimates (\ref{eq:lambda},\ref{eq:kappa}) using $\lambda=0.9, \kappa=0.1$ (as employed in the experiments) are not exact. 
}
Pattern capacity for {\bf auto-association} is significantly larger for auto-association than for hetero-association. The advantage of the Bayesian rule over BCPNN is also more significant for auto-association,
in particular for Willshaw patterns. 

%

\begin{figure*}[tp]
  \begin{center}
  \includegraphics[width=\linewidth]{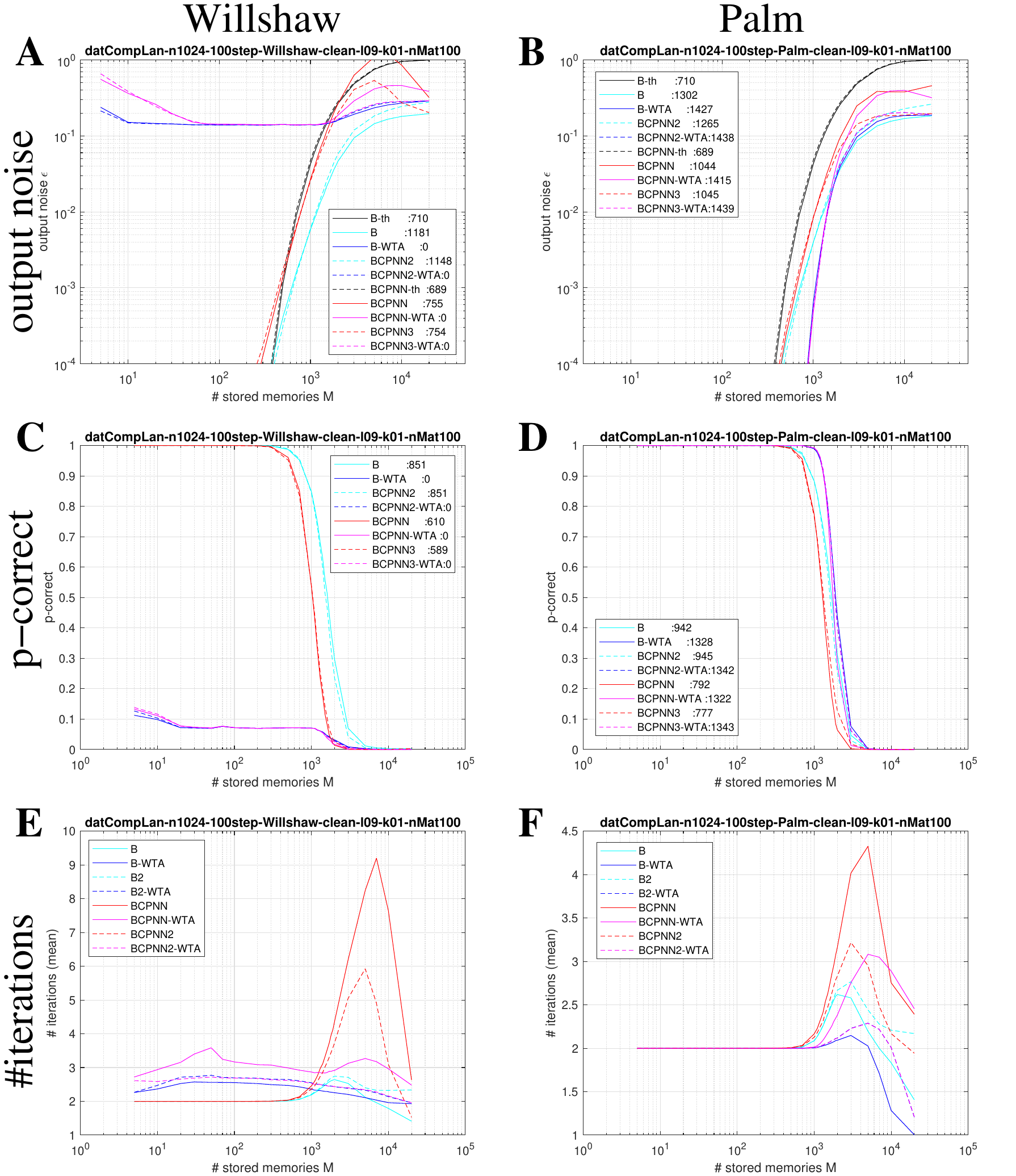}
  \end{center}
  \caption{\label{fig2:CompLan100step}
    Results for {\bf auto-association} with {\bf iterative retrieval} (max. 100 iterations) for Willshaw (left) and Palm patterns (right)
    for $n=1024$, $k=32$, $\lambda=0.9$, $\kappa=0.1$.
    {\bf A,B}: Output noise $\eps$ as function of stored memories $M$, similar to previous Fig.~\ref{fig1:ExpRepro2A}.
    {\bf C,D}: Fraction $p_{\mathrm{corr}}$ of correct retrieval outputs (corresponding to zero output noise $\eps^{\mu}=0$).
               Numbers in legends correspond to (interpolated) pattern capacity $M_{p_\mathrm{corr}}$ at $p_{\mathrm{corr}}=0.9$. 
    {\bf E,F}: Mean iteration number until convergence. 
  }
\end{figure*}

Then a {\bf third series of experiments} tested {\bf iterative retrieval} in the {\bf auto-associative networks},
allowing for a maximal number of 100 iterations per retrieval operation. 
Fig.~\ref{fig2:CompLan100step} shows corresponding results for network size $n=1024$, pattern activity $k=32(=\sqrt{n})$,
input noise $\lambda=0.9$, $\kappa=0.1$, and Willshaw (left panels) and Palm patterns (right panels), similar as in the previous Fig.~\ref{fig1:ExpRepro2A}.
The top panels (A,B) show {\bf output noise} $\eps$ as function of stored patterns $M$ (similar as in Fig.~\ref{fig1:ExpRepro2A}). Similar to previous experiments, the Bayesian rule yields best results,
at least for Willshaw patterns, but also for Palm patterns (but is slightly outperformed by the variants BCPNN-2/3 for WTA-retrieval). 
We also see that iterative retrieval significantly reduces output noise
compared to one-step retrieval. For example the Bayesian learning rule (``B'') reaches $M=1181$ at $\eps=0.01$ for Willshaw patterns with iterative retrieval,
compared to $M=1030$ for one-step retrieval (not shown)
\footnote{\label{footnote:interpolationeffects}
  Note that in Fig.~\ref{fig1:ExpRepro2A} we have estimated $M=1023$ at $\eps=0.01$ for auto-association of Willshaw patterns (instead of $M=1030$).
  This discrepancy can be attributed to {\bf random variations} (as patterns are random patterns), as well as to {\bf interpolation effects}: In Fig.~\ref{fig1:ExpRepro2A} we tested for
  $M=5, 10, 20, 30, 50 , 70, 100, 200, 300, 500, 700, 1000, 2000, 3000, 5000, 70000, 10000, 20000$,
  whereas in Fig.~\ref{fig2:CompLan100step} we tested additionally for $M=1100, 1200, 1300, 1400, 1500, 1600, 1700, 1800, 1900$ in order to have a higher resolution around the capacity limits.
  Due to the convex shape of the output noise curve, piecewise linear interpolation (as employed here) generally underestimates the true pattern capacity.
  And for the lower resolution of Fig.~\ref{fig1:ExpRepro2A}, this underestimation is stronger.
}, corresponding to an $\approx 15$ percent increase of pattern capacity.
To be more comparable with related works investigating the BCPNN rule \cite{Lansner/Ravichandran/Herman:2023,Lansner:2024:perscomm}, we tested also {\bf fractions of correct retrievals} $p_{\mathrm{corr}}$ (middle panels C,D).
Results are similar as for output noise: Again, the Bayesian rule gives best results for Willshaw patterns, whereas the variants BCPNN2/3 are best for Palm patterns.
The pattern capacities given at $p_{\mathrm{corr}}=0.9$ in the legends are a bit lower than at $\eps=0.01$, meaning that the former is a harder criterion than the latter.
The lower panels (E,F) show {\bf mean number of iterations until convergence}. In general, convergence is quick, where a small number ($<$10) is sufficient for convergence.
In particular, for low output noise (with few patterns stored), convergence is reached after one or two steps. Around the capacity limit the number of iterations increases (with increasing output noise).
There, the BCPNN-type rules appear to require significantly more iterations than the Bayesian rules (which may be due to higher output noise level at/above capacity limits,
preventing fast convergence). 

%

\begin{figure*}[ht]
  \begin{center}
  \includegraphics[width=\linewidth]{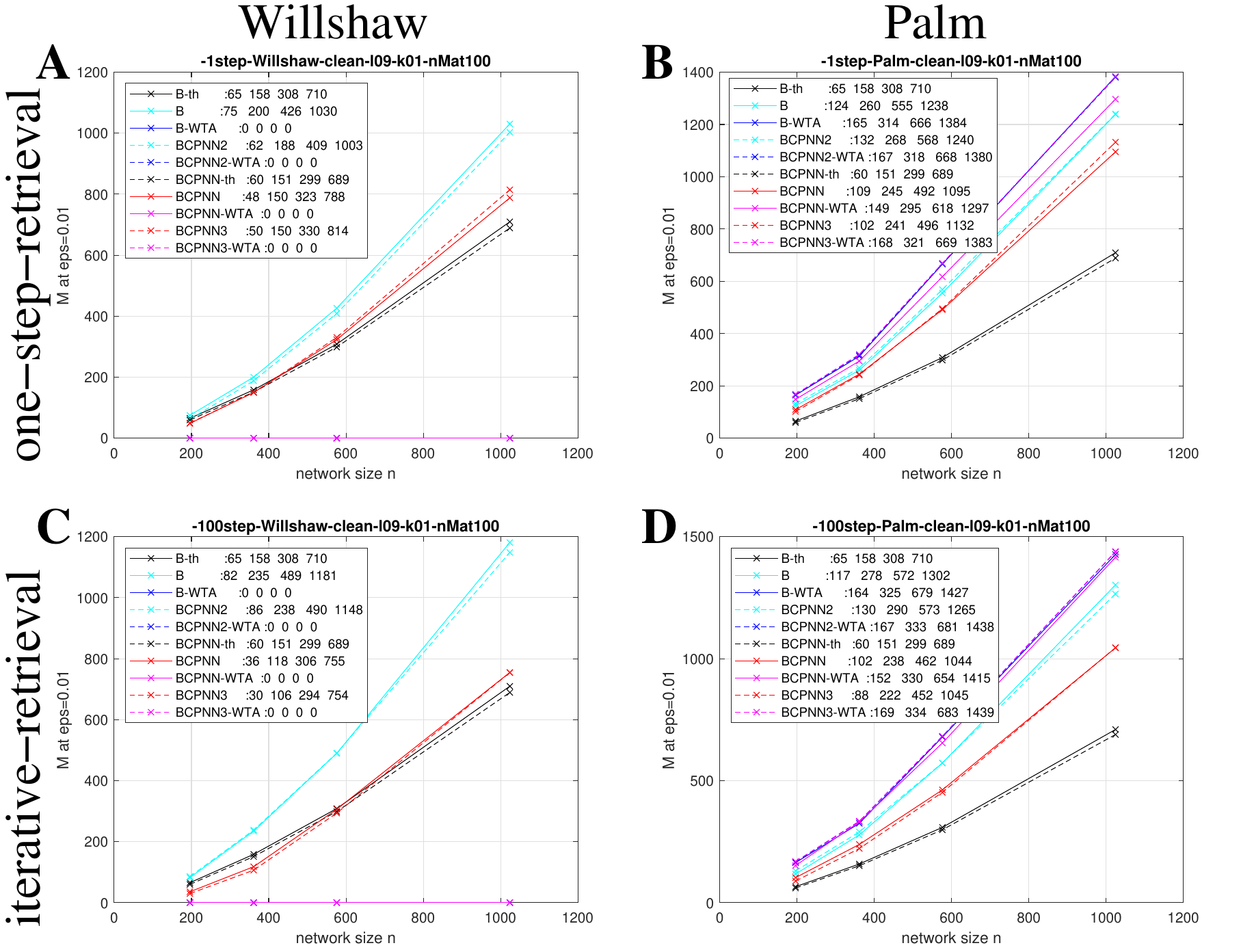}
  \end{center}
  \caption{\label{fig3:CompLan_M001}
    {\bf Pattern Capacity} $M_{\eps}$ {\bf at output noise level} $\eps=0.01$. Experimental setup is as in previous Fig.~\ref{fig2:CompLan100step}, but for
    different network sizes $n=196, 361, 576, 1024$ and pattern activity $k=\sqrt{n}=14, 19, 24, 32$. Numbers of legends correspond to (interpolated) pattern
    capacities for each $n$.
  }
\end{figure*}

\begin{figure*}[ht]
  \begin{center}
  \includegraphics[width=\linewidth]{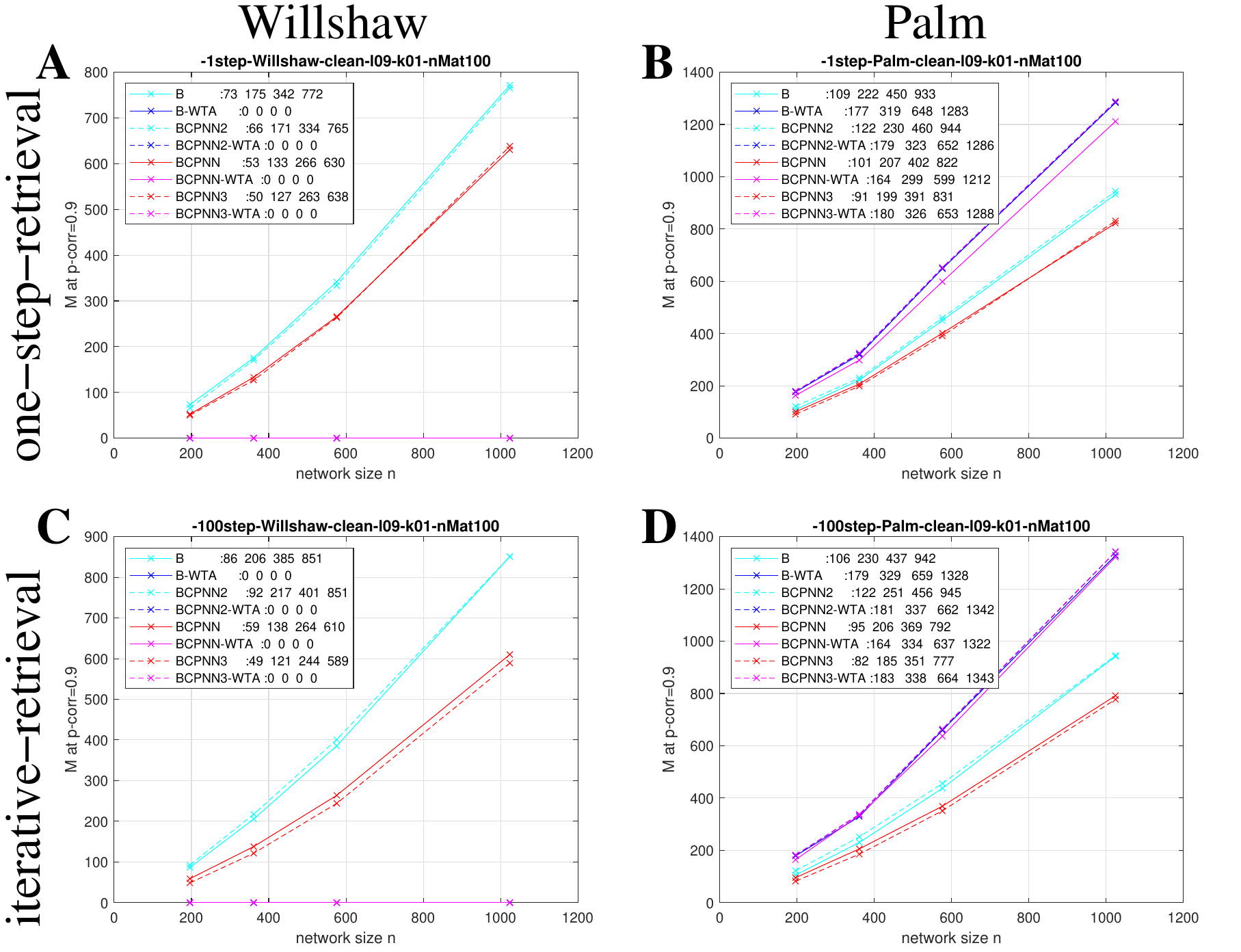}
  \end{center}
  \caption{\label{fig4:CompLan_M09}
    {\bf Pattern Capacity} $M_{p_{\mathrm{corr}}}$ {\bf at correctness level} $p_{\mathrm{corr}}=0.9$.
    Experimental setup is as in previous Figs.~\ref{fig2:CompLan100step},\ref{fig3:CompLan_M001} for
    network sizes $n=196, 361, 576, 1024$ and pattern activity $k=\sqrt{n}=14, 19, 24, 32$.
    Numbers of legends correspond to (interpolated) pattern capacities for each $n$.
  }
\end{figure*}

Next we repeated the experiments of Fig.~\ref{fig2:CompLan100step} for various network sizes
$n=196, 361, 576, 1024$ and corresponding pattern activities $k=\sqrt{n}=14, 19, 24, 32$. Fig.~\ref{fig3:CompLan_M001}
and Fig.~\ref{fig4:CompLan_M09}
show the corresponding {\bf maximal pattern capacities} at output noise level $\eps=0.01$ and at fraction of correct retrievals
$p_{\mathrm{corr}}=0.9$, respectively.
\footnote{\label{footnote:interpolationeffectsII}
  For the experiments we again tested pattern numbers $M$ with increased resolution at capacity limits taken from \cite{Lansner/Ravichandran/Herman:2023}. 
  That is, for all network sizes we tested $M=5, 10, 20, 30, 50, 70, 100, 200, 300, 500, 700, 1000, 2000, 3000, 5000, 7000, 10000, 20000$,
  where for network size $n=196$ we tested additionally $M=110, 120, 130, \ldots, 290$; for $n=361$ we tested additionally $M=125, 150, 175, \ldots, 475$;
  for $n=576$ we tested additionally $M=400, 450, \ldots, 850, 900, 950$; for $n=1024$ we tested additionally $M=1100, 1200, \ldots, 1900$ (as for Fig.~\ref{fig2:CompLan100step}).  
}
Again, we see that at least for Willshaw pattern (independent components), the Bayesian rule (``B'') is always optimal, both for one-step and iterative retrieval.
Even for Palm patterns, the Bayesian rule is always better than BCPNN, however, slightly outperformed by the WTA-variants of BCPNN2 and BCPNN3, most visibly
for iterative retrieval. Interestingly, while most learning rules have a {\bf significantly larger pattern capacity for iterative retrieval} compared to one-step retrieval (as expected),
the basic {\bf BCPNN rule (without WTA) becomes worse} for iterative retrieval. This may reflect the suboptimal threshold control of BCPNN without the $k$-WTA mechanism.

Thus, the {\bf intermediate results} so far can be summarized as follows:
\begin{itemize}
\item Under the theoretical assumptions of independent pattern components, independent input noise, and one-step-retrieval,
      the Bayesian rule (``B'') always provides the best results. 
\item Even for Willshaw-type patterns ($k$ of $n$), $k$-WTA, and iterative retrieval, the Bayesian rule performs still better than BCPNN,
      although sometimes (slightly) worse then variants BCPNN2/BCPNN3. This may be explained either as random fluctuations (as the deviations seem small)
      or by overfitting of the Bayesian rule to the special assumptions mentioned above.
\end{itemize}
In any case, so far we have not seen any anomalies that BCPNN performs (significantly) better than the optimal Bayesian rule.

\subsection{Reproducing and explaining the anomalies}\label{sec:experiments_results_anomalies}
So far not being able to reproduce the reported anomalies that BCPNN performs (significantly) better than the optimal Bayesian rule \cite{Lansner:2024:perscomm},
I started reconsidering iterative retrieval in auto-association: Obviously, the Bayesian learning rule will be optimal only in the first retrieval step,
when the input noise parameters $\lambda,\kappa$ are adjusted correctly to the actual input noise of the initial pattern $\tilde{u}$. After each iteration step,
the input noise for the next retrieval will be smaller, and thus the initial noise estimates $\lambda,\kappa$ are no longer correct.
As a consequence, the Bayesian rule employed so far (with the initial input noise estimates $\lambda,\kappa$) may not be optimal for iterative retrieval.
In a {\bf fourth series of experiments} I tested {\bf iterative retrieval} using {\bf different input noise estimates} $\lambda^{\mathrm{est}}/\kappa^{\mathrm{est}}$ for training the network,
deviating from the actual input noise $\lambda,\kappa$ in the initial query pattern $\tilde{u}$.
Table~\ref{tab1:M_pcorr} summarizes the results for $n=1024$, $k=32$ and $\lambda=0.9$, $\kappa=0.1$, showing maximal pattern capacity $M$ at
correctness $p_{\mathrm{corr}}\ge 0.9$ when training with various input noise estimates from $\lambda^{\mathrm{est}}=0.8, \kappa^{\mathrm{est}}=0.2$ to $\lambda^{\mathrm{est}}=1, \kappa^{\mathrm{est}}=0$.
Now we indeed observe the {\bf anomaly, that BCPNN-WTA can perform significantly better than the Bayesian rule for Palm patterns}. However, this requires the assumption of small input noise
(maximal $M=1433$ is achieved for $\lambda^{\mathrm{est}}=0.999, \kappa^{\mathrm{est}}=0.001$). For input noise estimates larger or smaller, pattern capacity BCPNN again drops to smaller values.
%
\begin{table}[ht]
\makebox[\linewidth]{\footnotesize
\begin{tabular}{l||cc|cc||cc|cc}
                                          & Palm            &                  &                  &                  & Willshaw        &                  &                  &           \\
                                          & BCPNN-WTA       &                  & B-WTA            &                  & BCPNN           &                  & B                &           \\
$\lambda^{\mathrm{est}}/\kappa^{\mathrm{est}}$   & 1 step          & 100 step         & 1 step           & 100 step         & 1 step          & 100 step         & 1 step           & 100 step  \\ \hline
  $0.8 / 0.2$                             & 1202            & 1260             & 1231             & 1275             & 416             & 391              & 584              & 535   \\
  $0.9 / 0.1$                             & \underline{1212}& 1322             & \underline{1283} & \underline{1328} & \underline{630} & 610              & \underline{772}  & 851    \\
  $0.99 / 0.01$                           & 1131            & 1418             & 1245             & 1327             & 373             & 1008             & 401              & \underline{1115} \\
  $0.999 / 0.001$                         & 1047            & \underline{1433} & 1167             & 1257             & 118             & \underline{1095} & 185              & 887  \\
  $0.9999 / 0.0001$                       & 986             & 1422             & 1115             & 1197             & 48              & 1022             & 117              & 725  \\
  $1-1e-5 / 1e-5$                         & 911             & 1386             & 1071             & 1153             & 25              & 539              & 88               & 533  \\
  $1-1e-6 / 1e-6$                         & 848             & 1357             & 1042             & 1131             & 14              & 186              & 76               & 333  \\
  $1      / 0$                            & 732             & 1049             &  783             & 1101             & 0               & 0                & 0                & 0 \\
\hline
\end{tabular}}
\caption{\label{tab1:M_pcorr}
  Maximal {\bf pattern capacity} $M$ satisfying $p_{\mathrm{corr}}\ge 0.9$ for auto-associative networks of size $n=1024$ with an average of $k=32$ active units per pattern and
  query noise $\lambda=0.9$, $\kappa=0.1$ with different noise estimates
  $\lambda^{\mathrm{est}}$, $\kappa^{\mathrm{est}}$ (first column) used for learning with the BCPNN or Bayesian (B) rule. Experimental setup as in previous Fig.~\ref{fig4:CompLan_M09}.
  Results correspond to 1-step-retrieval (columns 2,4,6,8) or iterative (max. 100-step) retrieval (columns 3,5,7,9). 
  Columns 2-5 correspond to ``Palm patterns'',
  where each pattern has exactly $k$ active units (and the query has $\mathrm{round}(\lambda k)$ correct and $\mathrm{round}(\kappa k)$ false active units).
  Columns 6-9 correspond to ``Willshaw patterns'', where each pattern component is active independently with probability $k/n$ (and queries components are modified independently
  according to (\ref{eq:lambda},\ref{eq:kappa})). 
  For Palm patterns we use $k$-winners-take-all retrieval (WTA), whereas for Willshaw patterns we use fixed firing threshold 0. Underlines correspond to maximum pattern capacity per column.
}
\end{table}
After communicating again with Anders Lansner \cite{Lansner:2024:perscommII}, I realized that, unlike to the original BCPNN rule (\ref{eq:wij_Bayesian_BCPNN_weights}) and
unlike to the precise implementation of \cite{Knoblauch:NeurComp2011}, app. A,
most {\bf recent implementations of BCPNN} \cite{Lansner/Holst:1996,Johansson/Lansner:2007,Martinez:BCPNN:2022,Lansner/Ravichandran/Herman:2023} actually employ the following {\bf numerically stabilized variant} 
      \begin{eqnarray}
          w_{ij}^{\mathrm{BCPNN}} & = & \log\frac{M_{11}'(ij)M}{M_1(j)M_1(i)}  \quad\quad\mbox{with}\quad M_{11}':=\max(M_{11},\eta \eps_s^2M) \label{eq:wij_Bayesian_BCPNN_weights_stable} 
      \end{eqnarray}
      where $\eta\eps_s>0$ avoids $\log(0)=\infty$ if $M_{11}=0$. Lansner and colleagues recommend $\eps_s:=\frac{1}{1+M}$ for $\eta=1$ to 
      optimize storage capacity (\cite{Lansner:2024:perscommII}; cf., \cite{Martinez:BCPNN:2022}, pp 52-54). 
      Obviously, the same numerical stabilization method can also be applied to the other Bayesian and BCPNN variants (\ref{eq:wij_Bayesian},\ref{eq:xj_Bayesian},\ref{eq:wij_noisy_BCPNN}-\ref{eq:wijBCPNNIII}),
      by simply replacing the critical synaptic counter variable $M_{11}$ by $M_{11}'$
      as in (\ref{eq:wij_Bayesian_BCPNN_weights_stable}).
      \footnote{\label{footnote:M11critical_BCPNN_poor_for_small_M}
        $M_{11}$ is the most critical counter variable as it is most likely to remain zero and cause infinite synaptic weights: For example, 
        for sparse random patterns with small $p:=k/n\ll 0$, it follows a binomial distribution $M_{11}\sim B(M,p^2)$ with parameters $M$ and $p^2$ (since $M$ patterns are stored, and for each pattern the
        chance of a pre-/postsynaptic coincidence is $p^2\ll 0$). By contrast, the other synaptic counters $M_{10},M_{01}\sim B(M,p(1-p))$ and $M_{00}\sim  B(M,(1-p)(1-p))$ are unlikely to remain zero, unless
        a very small number of pattern $M$ is stored. Correspondingly, as (\ref{eq:wij_Bayesian_BCPNN_weights_stable}) stabilizes only $M_{11}$, in the experiments of Table~\ref{tab2:M_pcorr_stab}
        BCPNN generally performed poor for $M<300$ (data not shown).
      }
      Table~\ref{tab2:M_pcorr_stab} shows corresponding results using different values for 
      scaling factor $\eta\in\{0.01,0.1,0.2,0.5,1,2,5,10,100\}$ of stabilization. 
%
\begin{table}[ht]
\makebox[\linewidth]{\footnotesize
\begin{tabular}{l||cc|cc||cc|cc}
                                               & Palm            &                  &                  &                  & Willshaw        &                  &                  &           \\
                                               & BCPNN-WTA       &                  & B-WTA            &                  & BCPNN           &                  & B                &           \\
$\lambda^{\mathrm{est}}/\kappa^{\mathrm{est}}/\eta$   & 1 step          & 100 step         & 1 step           & 100 step         & 1 step          & 100 step         & 1 step           & 100 step  \\ \hline
  $1 / 0 / 0.01$                               & 908             & 1383             & 0                & 0                & 0               & 740              & 0                & 0     \\
  $1 / 0 / 0.1$                                & 995             & 1415             & 0                & 0                & 0               & 1021             & 0                & 0     \\
  $1 / 0 / 0.2$                                & 1009            & 1420             & 0                & 0                & 0               & 1046             & 0                & 0     \\
  $1 / 0 / 0.5$                                & 1026            & 1434             & 0                & 0                & 0               & 1099             & 0                & 0     \\
  $1 / 0 / 1$                                  & 1042            & 1430             & 0                & 0                & 0               & \underline{1102} & 0                & 0      \\
  $1 / 0 / 2$                                  & 1059            & \underline{1439} & 0                & 0                & 0               & 1086             & 0                & 0    \\
  $1 / 0 / 5$                                  & 1102            & 1435             & 0                & 0                & 0               & 1064             & 0                & 0  \\
  $1 / 0 / 10$                                 & 1114            & 1434             & 0                & 0                & 561             & 1010             & 0                & 0     \\
  $1 / 0 / 100$                                & 1176            & 1353             & 0                & 0                & 0               & 0                & 0                & 0     \\
\hline
  $0.9 / 0.1 / 0.01$                           & 1210            & 1322             & 1273             & 1318             & 628             & 606              & 764              & 844     \\
  $0.9 / 0.1 / 0.1$                            & 1210            & 1326             & 1279             & 1325             & 626             & 614              & \underline{779}  & \underline{869} \\
  $0.9 / 0.1 / 0.2$                            & 1206            & 1331             & 1268             & 1327             & \underline{629} & 613              & 778              & 859     \\
  $0.9 / 0.1 / 0.5$                            & \underline{1212}& 1333             & \underline{1286} & 1331             & 627             & 608              & 777              & 843     \\
  $0.9 / 0.1 / 1$                              & \underline{1212}& 1328             & 1270             & 1324             & 622             & 608              & 773              & 855     \\
  $0.9 / 0.1 / 2$                              & 1209            & 1331             & 1273             & 1328             & 623             & 595              & 775              & 848     \\
  $0.9 / 0.1 / 5$                              & 1210            & 1332             & 1279             & \underline{1335} & 597             & 578              & 766              & 834     \\
  $0.9 / 0.1 / 10$                             & 1208            & 1321             & 1270             & 1318             & 565             & 532              & 766              & 801     \\
  $0.9 / 0.1 / 100$                            & 1175            & 1260             & 1224             & 1274             & 0               & 0                & 0                & 0     \\
\end{tabular}}
\caption{\label{tab2:M_pcorr_stab}
  Maximal {\bf pattern capacity} $M$ satisfying $p_{\mathrm{corr}}\ge 0.9$ for auto-associative networks of size $n=1024$ with an average of $k=32$ active units per pattern and
  query noise $\lambda=0.9$, $\kappa=0.1$ using zero noise-estimates $\lambda^{\mathrm{est}}=1, \kappa^{\mathrm{est}}=0$ ({\bf upper part})
  or correct noise-estimates $\lambda^{\mathrm{est}}=0.9, \kappa^{\mathrm{est}}=0.1$ ({\bf lower part})
  for {\bf numerical stabilized} $M_{11}'$ as in (\ref{eq:wij_Bayesian_BCPNN_weights_stable})
  for different scaling factors $\eta\in\{0.01,0.1,0.2,0.5,1,2,5,10,100\}$. The case $\eta=1$ corresponds to stabilization as
  considered optimal in prior studies \cite{Lansner:2024:perscommII,Martinez:BCPNN:2022,Lansner/Ravichandran/Herman:2023}.
  Otherwise, the experimental setting was as in the previous Table~\ref{tab1:M_pcorr}.
}
\end{table}
We see that for zero-noise-estimation ($\lambda^{\mathrm{est}}=1, \kappa^{\mathrm{est}}=0$; upper part of the table), BCPNN achieves the maximal pattern capacity $M=1439$ for Palm patterns with $\eta=2$,
where similarly high values occur also in the range $\eta\in[0.5;10]$, which is consistent with the reports of Lansner et al. for $\eta=1$ \cite{Lansner:2024:perscomm,Lansner/Ravichandran/Herman:2023}.
For Willshaw patterns the pattern capacity of BCPNN is significant lower ($M=1102$ for $\eta=1$). Due to the wrong noise estimates, here the Bayesian learning rule never exceeds the 90\% correct thresholds (always $M=0$).
For correct initial noise estimates ($\lambda^{\mathrm{est}}=0.9, \kappa^{\mathrm{est}}=0.1$; lower part of the table), both BCPNN and the Bayesian rule with stabilization perform similar as in Table~\ref{tab1:M_pcorr}
without stabilization. It can also be seen that the Bayesian learning rule is more sensitive to reasonable noise estimates $\lambda^{\mathrm{est}}, \kappa^{\mathrm{est}}$ than BCPNN. Although stabilized BCNN reaches
very high maximal capacities, it performs quite bad for few stored memories (e.g., for $M<300$ the fraction of correct retrievals was close to zero; data not shown;
cf. previous footnote~\ref{footnote:M11critical_BCPNN_poor_for_small_M}).
We may summarize the {\bf intermediate results} as follows:
\begin{itemize}
\item Table~\ref{tab2:M_pcorr_stab} suggests that the anomalous high storage capacity of the BCPNN model \cite{Lansner:2024:perscomm,Lansner/Ravichandran/Herman:2023} seems to be mainly due to optimizing the regularization/stabilization parameter
      $\eps_s$ (for $\eta=1$) in the recent BCPNN learning rule (\ref{eq:wij_Bayesian_BCPNN_weights_stable}).
\item In fact, for sparse random patterns the synaptic counter $M_{11}\sim B(M,p^2)$ with $p:=\frac{k}{n}=\frac{\sqrt{n}}{n}=\frac{1}{\sqrt{n}}$ has still a high change of being zero:
      As $M_{11}$ is binomially distributed, for $k=\sqrt{n}$ the probability $p[M_{11}=0]$ that $M_{11}$ is zero is still $(1-p^2)^M>0$ (although $p[M_{11}=0]\rightarrow 0$ asymptotically;
      see ``dense potentiation'' regime in \cite{Knoblauch/Palm/Sommer:NeurComp2010}, sect.~3.4). 
      For example, for $n=1024$, $k=32$, $M=1433$ (corresponding to maximal capacity in Tab.~\ref{tab1:M_pcorr}), it is still $p[M_{11}=0]=(1-(32/1024)^2)^{1433}\approx 0.24$.
      Thus, about a quarter of all synapses will have $M_{11}=0$ and would produce (negative) infinite weights according to the ``pure'' BCPNN rule (\ref{eq:wij_Bayesian_BCPNN_weights})
      (or similarly according to the Bayesian rule (\ref{eq:wij_Bayesian}) or BCPNN rule (\ref{eq:wij_noisy_BCPNN}) with zero add noise estimates $p_{01}=0$). 
\item So tuning the regularization/stabilization parameter $\eps$ will have a huge impact on the resulting network, since it rescales a large fraction of the synapses from $-\infty$ to a range of
      finite values that strongly depends on the precise value of $\eps_s$.
      Some unpublished theoretical works from Anders Lansner's group have
      suggested optimal values $\eps=\frac{1}{1+M}$ such that $\eps$ can be interpreted as the ``smallest probability measurable in the system''
      when storing $M$ patterns (\cite{Lansner:2024:perscommII};\cite{Martinez:BCPNN:2022}, pp 52-54). 
\item By contrast, the ``pure'' BCPNN model (\ref{eq:wij_Bayesian_BCPNN_weights}),
      even if implemented in an exact way considering infinite weights (\cite{Knoblauch:NeurComp2011}, app.A) 
      performs much worse, both for the original variant (not considering input noise) and the generalized model including realistic noise estimates (of the inputs) using the generalized rule (\ref{eq:wij_noisy_BCPNN}).
      In particular, Table~\ref{tab1:M_pcorr} shows that the Bayesian rule remains superior over BCPNN in these cases, and there are no anomalies. 
\item However, according Table~\ref{tab1:M_pcorr} the anomalies occur as well if we use input noise estimates $\lambda^{\mathrm{est}},\kappa^{\mathrm{est}}$ lower than the actual input noise.
      For example, if for $\lambda=0.9$,$\kappa=0.1$ we
      estimate $\lambda^{\mathrm{est}}=0.999$, $\kappa^{\mathrm{est}}=0.001$, the BCPNN model reaches pattern capacity $M=1433$ (for Palm patterns), significantly above the pattern capacity of the ``optimal''
      Bayesian rule (only $M=1328$ for $\lambda^{\mathrm{est}}=0.9$, $\kappa^{\mathrm{est}}=0.1$).
\item As here the maximum pattern capacity of the BCPNN rule is similar as previously reported \cite{Lansner:2024:perscomm,Lansner/Ravichandran/Herman:2023}, this  
      suggests that the stabilization method (\ref{eq:wij_Bayesian_BCPNN_weights_stable}) has a similar effect as using small error estimates corresponding to the employed quality criterion (note that the optimal estimates
      $\lambda^{\mathrm{est}}=0.999$, $\kappa^{\mathrm{est}}=0.001$ are for $n=1024$ indeed close to the zero error criterion ($p_{\mathrm{corr}}\ge 0.9$) used for determining pattern capacity in Tab.~\ref{tab1:M_pcorr}). 
      However, more theoretical work may be necessary to establish a more quantitative connection.
\item In conclusion, the optimal {\bf Bayesian rule seems to be much more sensitive to correct input noise estimates} $\lambda^{\mathrm{est}},\kappa^{\mathrm{est}}$ than BCPNN.
      As during iterative retrieval, the input noise will diminish from large values to zero ($\lambda^{\mathrm{est}}\rightarrow 1$, $\kappa^{\mathrm{est}}\rightarrow 0$), it is logically impossible to maintain
      correct input noise estimates over all retrieval steps, leading to suboptimal retrieval results.
      \footnote{
        The {\bf degrading effect of wrong / constant input noise estimates} can be seen, for example, in Table~\ref{tab1:M_pcorr}, \ref{tab2:M_pcorr_stab} for the Willshaw patterns,
        where 100-step iterative retrieval often performs worse than 1-step retrieval (in particular, if the noise estimates approximately match the actual input noise in the first iteration step).
      }
      Thus, due to its greater robustness against wrong input noise estimates,
      {\bf BCPNN can perform better for constant input noise estimates} than the Bayesian learning rule.
\end{itemize}
Overall, these results give a sufficient {\bf explanation of the anomaly} that the stabilized BCPNN rule (\ref{eq:wij_Bayesian_BCPNN_weights_stable}) can sometimes (for Palm patterns, iterative retrieval, WTA)
perform better than the ``optimal'' Bayesian learning rule. Correspondingly, we can reject the alternative hypothesis, that the implementation of the Bayesian rule would be erroneous.
This reasoning suggests also that the performance of the Bayesian learning could be improved if input noise estimates would be adapted during iterative retrieval as explored in the next section.

\subsection{Improving Bayesian learning by Adaptive Noise Estimation (ANE) over iterative retrieval}\label{sec:experiments_results_ANE}
Estimating input noise by parameters $\lambda^{\mathrm{est}}$, $\kappa^{\mathrm{est}}$ will lead to optimal retrieval in a one-step-retrieval scenario, but obviously not
for iterative retrieval. This is because after each iteration the noise becomes smaller, and the initial prior estimates $\lambda^{\mathrm{est}}$, $\kappa^{\mathrm{est}}$ will become invalid.
This reasoning immediately suggests that the performance of the Bayesian rule may improve for iterative retrieval if the noise estimates $\lambda^{\mathrm{est}}(t)$, $\kappa^{\mathrm{est}}(t)$ could be adapted
over iterative retrieval time $t$ ({\bf Adaptive Noise Estimation; ANE}). Thus, it should be optimal to start with high input noise estimates $\lambda^{\mathrm{est}}(1)=\lambda$, $\kappa^{\mathrm{est}}(1)=\kappa$ corresponding to the true input query noise
in the first step ($t=1$), and then decreasing the estimates toward zero ($\lambda^{\mathrm{est}}(t)\rightarrow 1$, $\kappa^{\mathrm{est}}(t)\rightarrow 0$) in the further iteration steps $t=2,3,\ldots$.

Table~\ref{tab3:ANE} shows results from a {\bf first attempt}, where I have adapted the noise estimates $\lambda^{\mathrm{est}}(t)$ and $\kappa^{\mathrm{est}}(t)$ of the optimal Bayesian rule
in each step $t>1$ to the measured output noise $\lambda^{\mathrm{out}}(t-1)$ and $\kappa^{\mathrm{out}}(t-1)$ of the previous step $t-1$. Tracing the output noise over iteration steps reveals that 
only the first two or three iteration steps are relevant, whereas further iteration steps have only minor effects on retrieval quality, even if convergence takes longer.
ANE requires the noise to be estimated for a particular reference number of stored patterns $M$. As I hoped to significantly improve pattern capacity by ANE,  I chose reference
$M$ moderately larger than the maximal pattern capacity in Table~\ref{tab1:M_pcorr} ($M=1400$ for Palm patterns and $M=1200$ for Willshaw pattern; compared to
maximum $M=1328$ for Palm/B-WTA and $M=1115$ for Willshaw/B in Table~\ref{tab1:M_pcorr}). However, for this setting, maximal $p_{\mathrm{corr}}$ remained below the threshold of 90 percent
($p_{\mathrm{corr}}=0.8781$ at iteration step 5 for Palm/B-WTA; $p_{\mathrm{corr}}=0.8235$ at iteration step 10 for Willshaw/B).

%
\begin{table}[ht]
\makebox[\linewidth]{\footnotesize
\begin{tabular}{c||c|cccc}
Bayes (B-WTA), Palm, $M=1400$  & \\
step $t$                  & $\lambda^{\mathrm{est}}/\kappa^{\mathrm{est}}$   & $\eps$  & $p_{\mathrm{corr}}$    &   $f_{10}/f_{01}$   &  $\lambda^{\mathrm{out}}/\kappa^{\mathrm{out}}$ \\ \hline
1                         & 0.90625 / 0.0937500                       & 0.011060& 0.8263             &   0.1770/0.1770     & 0.99447 / 0.0055312                      \\
2                         & 0.99447 / 0.0055312                       & 0.008044& 0.8737             &   0.1287/0.1287     & 0.99598 / 0.0040219                    \\
3                         & 0.99598 / 0.0040219                       & 0.007906& 0.8758             &   0.1265/0.1265     & 0.99605 / 0.0039531                    \\
4                         & 0.99605 / 0.0039531                       & 0.008025& 0.8739             &   0.1284/0.1285     & 0.99599 / 0.0040125                    \\
5                         & 0.99599 / 0.0040125                       & 0.007769& \underline{0.8781} &   0.1243/0.1243     & 0.99612 / 0.0038844                    \\
6                         & 0.99612 / 0.0038844                       & 0.007956& 0.8758             &   0.1273/0.1273     & 0.99602 / 0.0039781                    \\
7                         & 0.99602 / 0.0039781                       & 0.007969& 0.8742             &   0.1275/0.1275     & 0.99602 / 0.0039844                    \\
8                         & 0.99602 / 0.0039844                       & 0.008319& 0.8697             &   0.1331/0.1331     & 0.99584 / 0.0041594                    \\
9                         & 0.99584 / 0.0041594                       & 0.007844& 0.8763             &   0.1255/0.1255     & 0.99608 / 0.0039219                    \\
10                        & 0.99608 / 0.0039219                       & 0.008100& 0.8733             &   0.1296/0.1296     & 0.99595 / 0.0040500                    \\
100                       & 0.99608 / 0.0039219                       & 0.008288& 0.8697             &   0.1326/0.1326     & 0.99586 / 0.0041437                    \\
\hline
Bayes (B), Willshaw, $M=1200$  & \\
step $t$                  & $\lambda^{\mathrm{est}}/\kappa^{\mathrm{est}}$   & $\eps$   & $p_{\mathrm{corr}}$    &   $f_{10}/f_{01}$   &  $\lambda^{\mathrm{out}}/\kappa^{\mathrm{out}}$ \\ \hline
1                         & 0.90000 / 0.1000000                       & 0.015890 & 0.6561             &   0.3059/0.2025   &   0.99044 / 0.0063280     \\
2                         & 0.99044 / 0.0063280                       & 0.008003 & 0.8037             &   0.1143/0.1418   &   0.99643 / 0.0044310     \\
3                         & 0.99643 / 0.0044310                       & 0.007363 & 0.8187             &   0.1120/0.1236   &   0.99650 / 0.0038625     \\
4                         & 0.99650 / 0.0038625                       & 0.007275 & 0.8223             &   0.1051/0.1277   &   0.99672 / 0.0039906     \\
5                         & 0.99672 / 0.0039906                       & 0.007372 & 0.8183             &   0.1040/0.1319   &   0.99675 / 0.0041219     \\
6                         & 0.99675 / 0.0041219                       & 0.007816 & 0.8100             &   0.1151/0.1350   &   0.99640 / 0.0042188     \\
7                         & 0.99640 / 0.0042188                       & 0.007338 & 0.8185             &   0.1080/0.1268   &   0.99662 / 0.0039625     \\
8                         & 0.99662 / 0.0039625                       & 0.007450 & 0.8180             &   0.1086/0.1298   &   0.99661 / 0.0040562     \\
9                         & 0.99661 / 0.0040562                       & 0.007510 & 0.8141             &   0.1088/0.1314   &   0.99660 / 0.0041062     \\
10                        & 0.99660 / 0.0041062                       & 0.007222 & \underline{0.8235} &   0.1023/0.1288   &   0.99680 / 0.0040250     \\
100                       & 0.99660 / 0.0041062                       & 0.007434 & 0.8198             &   0.1089/0.1290   &   0.99660 / 0.0040313     \\
\hline
\end{tabular}}
\caption{\label{tab3:ANE}
  Results for the {\bf Bayesian learning rule with adaptive noise estimation (ANE)}. {\bf Upper part:} ANE for each step $t$ of iterative (WTA-) retrieval when storing $M=1400$ Palm patterns ($n=1024, k=32$).
  $\lambda^{\mathrm{est}}/\kappa^{\mathrm{est}}$ are estimations of input noise (as before), $\eps$ is output noise (mean errors normalized to mean activity $k$, $p_{\mathrm{corr}}$ is fraction of correct retrievals,
  $f_{10}/f_{01}$ are mean number of false negative/false positive components, and correspondingly $\lambda^{\mathrm{out}}:=1-f_{10}$, $\kappa^{\mathrm{out}}:=f_{01}/k$ code mean output noise measurements.
  ANE means here to use $\lambda^{\mathrm{est}}(t+1):=\lambda^{\mathrm{out}}(t)$ and $\kappa^{\mathrm{est}}(t+1):=\kappa^{\mathrm{out}}(t)$ for $t>1$. Unlike to the previous experiments, here we
  used $\lambda^{\mathrm{est}}(1)=0.90625>0.9$, $\kappa^{\mathrm{est}}(1)=0.0.09375<0.1$ to account for the rounding effects for Palm patterns (see footnote~\ref{footnote:rounding4Palmpatterns}).
  For iteration steps $t=11,\ldots,100$ we used $\lambda^{\mathrm{est}}(t):=\lambda^{\mathrm{out}}(10)$ and $\kappa^{\mathrm{est}}(t):=\kappa^{\mathrm{out}}(10)$
  {\bf Lower part:} Corresponding ANE for iterative retrieval when storing $M=1200$ Willshaw patterns.
}
\end{table}

Table~\ref{tab4:ANE_maxM} shows the corresponding (interpolated) maximal pattern capacity $M$ at $p_{\mathrm{corr}}=0.9$ as function of iteration step.
\footnote{
  While the experiments of Table~\ref{tab3:ANE} estimated noise (for ANE) at fixed pattern numbers $M=1400$ (Palm/B-WTA) and $M=1200$ (Willshaw/B), they
  evaluated also networks with other pattern numbers $M$ around these reference values, with the same or higher resolution as in the previous experiments
  ($M=1100,1200,\ldots,1700$ for Palm/B-WTA and $M=900,1000,\ldots,1500$ for Willshaw/B; 
  cf., footnotes~\ref{footnote:interpolationeffects},\ref{footnote:interpolationeffectsII}). Thus, it was possible (by interpolation) to estimate maximal pattern capacity at each iteration step,
  as in the previous experiments.
}
Similar as $p_{\mathrm{corr}}$, pattern capacity quickly reaches maximum values after a few iteration steps. {\bf ANE always improved pattern capacity} compared to the control cases (without any ANE).
However, the {\bf improvements are moderate}: For Palm/B-WTA the increase from $M=1330$ (control) to $M=1355$ is less than 2 percent. By contrast, for Willshaw/B the increase
from $M=851$ (control/see Tab.~\ref{tab1:M_pcorr}) to $M=1056$ is almost 25 percent, however, still not reaching the maximum $M=1115$ for constant low noise estimates in Table~\ref{tab1:M_pcorr}.
For Palm/BCPNN-WTA and Willshaw/BCPNN the increases $M=1332\rightarrow 1470$ and $M=1021\rightarrow 1056$ are about 10 percent and 3 percent where, surprisingly, the former (but not the latter)
exceeds both maxima of Tab.~\ref{tab1:M_pcorr} and Tab.~\ref{tab2:M_pcorr_stab} for constant low noise estimates ($M=1433$) and stabilized BCPNN ($M=1439$). 

%
\begin{table}[ht]
\makebox[\linewidth]{\footnotesize
\begin{tabular}{c||cc|cc|cc|cc}
Palm patterns,\\ ANE for $M=1400$  & \\
step $t$                  & B-WTA           & control         & BCPNN-WTA       & control         & BCPNN2-WTA      & control         & BCPNN3-WTA      & control  \\ \hline
1                         & 1272            & 1278            & 1212            & 1211            & 1279            & 1283            & 1281            & 1286     \\
2                         & 1353            & 1326            & 1449            & 1329            & 1396            & 1343            & 1398            & 1344     \\
3                         & 1348            & 1326            & 1459            & 1329            & 1405            & 1343            & 1407            & 1344     \\
4                         & 1348            & \underline{1330}& 1463            & \underline{1332}& 1406            & \underline{1348}& 1408            & \underline{1351} \\
5                         & \underline{1355}& 1325            & 1468            & \underline{1332}& 1407            & 1340            & 1409            & 1341     \\
6                         & 1354            & 1323            & \underline{1470}& 1321            & \underline{1413}& 1340            & \underline{1412}& 1343     \\
7                         & 1349            & \underline{1330}& 1465            & 1328            & 1411            & 1346            & 1410            & 1347     \\
8                         & 1348            & 1325            & 1463            & \underline{1332}& 1401            & 1340            & 1402            & 1341     \\
9                         & 1352            & 1328            & 1463            & \underline{1332}& 1410            & 1341            & 1411            & 1343     \\
10                        & 1348            & 1328            & 1457            & 1329            & 1405            & 1341            & 1408            & 1344     \\
100                       & 1347            & 1323            & 1462            & 1331            & 1406            & 1343            & 1407            & 1346     \\
\hline
Willshaw patterns,\\ ANE for $M=1200$  & \\
step $t$                  & B               & control& BCPNN           & control& BCPNN2         & control       & BCPNN3  & control  \\ \hline
1                         & $<$900          & $<$900 & $<$900          & $<$900 &   $<$900       & $<$900        &  $<$900 & $<$900         \\
2                         & 1028            & $<$900 & 955             & $<$900 &   1065         & $<$900        &  $<$900 & $<$900         \\
3                         & 1039            & $<$900 & 994             & $<$900 &   1073         & $<$900        &  $<$900 & $<$900         \\
4                         & 1047            & $<$900 & 1013            & $<$900 &   1091         & $<$900        &  $<$900 & $<$900         \\
5                         & 1048            & $<$900 & 1010            & $<$900 &   1086         & $<$900        &  $<$900 & $<$900         \\
6                         & 1050            & $<$900 & \underline{1021}& $<$900 &   1085         & $<$900        &  $<$900 & $<$900         \\
7                         & 1054            & $<$900 & 1010            & $<$900 &\underline{1096}& $<$900        &  $<$900 & $<$900         \\
8                         & 1053            & $<$900 & 1015            & $<$900 &   1084         & $<$900        &  $<$900 & $<$900         \\
9                         & 1043            & $<$900 & 1007            & $<$900 &   1090         & $<$900        &  $<$900 & $<$900         \\
10                        & \underline{1056}& $<$900 & 1017            & $<$900 &   1093         & $<$900        &  $<$900 & $<$900         \\
100                       & 1041            & $<$900 & 1008            & $<$900 &   1080         & $<$900        &  $<$900 & $<$900         \\
\hline
\end{tabular}}
\caption{\label{tab4:ANE_maxM}
  Estimations of {\bf maximal pattern capacity} at $p_{\mathrm{corr}}=0.9$ {\bf for ANE} corresponding to the experiments of Table~\ref{tab3:ANE} for different learning rules (B=Bayes, BCPNN, BCPNN2, BCPNN3).
  {\bf Control values} correspond to experiments without ANE (that is, $\lambda^{\mathrm{est}}(t):=\lambda^{\mathrm{est}}(1)$ and $\kappa^{\mathrm{est}}(t):=\kappa^{\mathrm{est}}(1)$;
  compare also to Table~\ref{tab1:M_pcorr} for $\lambda^{\mathrm{est}}/\kappa^{\mathrm{est}}=0.9/0.1$). 
  Note that, for $t>1$, ANE always improves pattern capacity compared to the control experiments.
}
\end{table}

Thus, from these experiments, we may draw the following {\bf intermediate conclusions}:
\begin{itemize}
\item During iterative retrieval, maximum performance is typically reached within a few iteration steps (e.g., 3-5). The major gain is actually achieved in the 2nd retrieval step, whereas the
      remaining steps provide only minor improvements, if any.
\item Adaptive noise estimation (ANE) generally improves retrieval quality during iterative retrieval.
\item The improvements for the Bayesian learning rule are moderate, and do not exceed pattern capacity already reached in the previous experiments (for constant low noise estimates or for stabilized learning rules).
\item Surprisingly, BCPNN with ANE ($M=1470$) can exceed its previous maximum pattern capacities. 
\item Again, the reason for the high capacity of BCPNN seems related to its simplicity and, correspondingly, to its higher robustness against wrong noise estimates: In fact,
      although ANE aims to estimate the correct noise level in each step, the estimates $\lambda^{\mathrm{est}}(t)$ and $\kappa^{\mathrm{est}}(t)$ in the current experiments are only averages over many networks and
      retrievals (and units). By contrast, for individual networks and retrievals, the true values of $\lambda^{\mathrm{est}}(t)$ and $\kappa^{\mathrm{est}}(t)$ may vary considerably and, thus, explain
      the inferior performance of the Bayesian learning rule.
      \footnote{
        Note that the noise estimates like (\ref{eq:lambda},\ref{eq:kappa}) and error probabilities (\ref{eq:p01a},\ref{eq:p10a}) are synapse-dependent quantities (that is, they depend on $ij$).
        Applying simplified noise estimators like (\ref{eq:error_prob_indep_of_a}) (neglecting the dependencies on $i$ and $j$)
        is strictly valid only in the first retrieval step (when noise is generated in a controlled way by random number generation),
        whereas in ensuing retrieval steps the noise comes from the network and may vary considerably among different neuron pairs $ij$.
        Also in more realistic application scenarios (or in brain networks) the noise will depend on $ij$ already in retrieval step 1. 
      }
\end{itemize}
\subsection{ANE with core- and halo-retrieval for Palm patterns}\label{sec:experiments_results_ANE_corehalo_Willshaw}
In order to find conditions where the Bayesian learning rule is superior, the last item suggests to {\bf ``design'' retrieval conditions} where the {\bf query noise becomes predictable}.
For example, if we do $K$-WTA retrieval for Palm patterns with $K:=\alpha k$ and sufficiently small
$\alpha<1$, then all activated output units will be ``correct'' with high probability (since only the most strongly excited units get activated),
and we will obtain only miss noise in the retrieval outputs, that is, component errors will be
\begin{align}
  f_{01}=0\ \mbox{and}\ f_{10}=k-\alpha k    \quad\quad\mbox{or equivalently}\quad\quad \lambda^{\mathrm{est}}:=1-f_{10}/k=\alpha\ \mbox{and}\  \kappa^{\mathrm{est}}:=\frac{f_{01}}{k}=0   \label{eq:predictable_noise_est}
\end{align}
with high probability.
\footnote{
   Here $f_{01}$ and $f_{10}$ are numbers of false positive and false negative component errors in a retrieval output pattern, similar as in Table~\ref{tab3:ANE}. 
}
Thus, in a {\bf second attempt} of ANE we may employ the following fixed {\bf core-retrieval}
\footnote{
   {\bf ``Core-retrieval''} means that a subset of the ``true'' units (the ``core'' of a cell assembly) gets activated (see \cite{KnoblauchPalm:NeurComp2020}, Fig.~1). 
}
schedule for Palm patterns,
\begin{align}
  &\mbox{Iteration step 1:}           &\lambda^{\mathrm{est}}&:=\lambda, \quad &\kappa^{\mathrm{est}}&:=\kappa, \quad &\alpha&<1 \nonumber\\
  &\mbox{Iteration step 2:}           &\lambda^{\mathrm{est}}&:=\alpha , \quad &\kappa^{\mathrm{est}}&:=0,      \quad &\alpha&:=1 \nonumber\\
  &\mbox{Iteration steps 3,4,\ldots:} &\lambda^{\mathrm{est}}&:=1-\beta, \quad &\kappa^{\mathrm{est}}&:=\beta,  \quad &\alpha&:=1 \label{eq:coreretrieval}
\end{align}
where $\beta\ge 0$ is a small positive constant (e.g., $\beta:=0.01$) to enforce low error estimates in the last iteration steps (that proved to be beneficial in Tab.~\ref{tab1:M_pcorr}). 
A corresponding alternative would be to do $K$-WTA retrieval for Palm patterns with $K:=\alpha k$ and $\alpha>1$: If $\alpha$ is large enough then all ``correct'' units get activated plus perhaps some spurious units.
In analogy to (\ref{eq:coreretrieval}) this leads to the following fixed {\bf halo-retrieval}
\footnote{
   {\bf ``Halo-retrieval''} means that a superset of the ``true'' units (the ``halo'' of a cell assembly) gets activated (see \cite{KnoblauchPalm:NeurComp2020}, Fig.~1). 
}
schedule for Palm patterns,
\begin{align}
  &\mbox{Iteration step 1:}           &\lambda^{\mathrm{est}}&:=\lambda, \quad &\kappa^{\mathrm{est}}&:=\kappa, \quad &\alpha&>1 \nonumber\\
  &\mbox{Iteration step 2:}           &\lambda^{\mathrm{est}}&:=0 , \quad &\kappa^{\mathrm{est}}&:=\alpha-1,      \quad &\alpha&:=1 \nonumber\\
  &\mbox{Iteration steps 3,4,\ldots:} &\lambda^{\mathrm{est}}&:=1-\beta, \quad &\kappa^{\mathrm{est}}&:=\beta,  \quad &\alpha&:=1 \label{eq:haloretrieval}
\end{align}
Table~\ref{tab5:ANE_maxM_corehalo} shows results from experiments testing core- and halo-retrieval for our standard network ($n=1024,k=32$, Palm patterns, $\lambda=0.9,\kappa=0.1$)
for different values of $\alpha$ and $\beta$ after $t=5$ iteration steps (as this seemed to be a good choice in the previous experiments). In a first series of experiments we used $\beta=0.01$ (see above) and
optimized $\alpha\in\{0.78125, 0.8125, 0.84375, 0.875, 0.90625, 0.9375, 0.96875\}$ for core-retrieval (corresponding to $\alpha k=25,26,\ldots,31$ active units after the first iteration step)
and $\alpha\in\{1.0312, 1.0625, 1.0938, 1.125, 1.1562, 1.1875, 1.2188\}$ for halo-retrieval (corresponding to $\alpha k=33,34,\ldots,39$ active units after the first iteration step).
Then, in a second series of experiments, we tested different values of $\beta\in\{0.1,0.01,0.001,0.0001,0.00001,0\}$ for the two best values of $\alpha$.
The results of the experiments can be summarized as follows:
\begin{itemize}
\item For all learning rules best results are obtained if $\alpha$ is {\bf as close to 1 as possible}: Thus, for core-retrieval $\alpha=0.96875$ is optimal, activating in the first retrieval step
  $31$ of the $k=32$ neurons of a cell assembly. Similarly, for halo-retrieval $\alpha=1.0312$ is optimal, activating the cell assembly and one addition ``spurious'' neuron.
\item For $\beta$ it is beneficial to choose a {\bf reasonable small value} like $\beta=0.001$. This is consistent with the results of Tab.~\ref{tab1:M_pcorr}, whereas here
  we may choose $\beta$ even smaller (but not zero) in accordance with the low expected noise levels in the last retrieval steps.  
\item Strikingly, {\bf core-retrieval} significantly {\bf improves the capacity of all learning rules}. 
  In particular, now the {\bf Bayesian learning rule (B-WTA) achieves highest capacity} with $M=1593$ for $\alpha=0.96875, \beta=0.001$, corresponding to
  a {\bf 20 percent increase} compared to fixed noise estimates ($M=1328$ in Tab.~\ref{tab1:M_pcorr}). In comparison, the increase $M=1433\rightarrow 1517$ for BCPNN-WTA is still significant ($>5$ percent),
  but clearly below the Bayesian rule. The other rules BCPNN2/3-WTA are in between ($M\rightarrow 1559/55$). 
\item By contrast, the improvements of {\bf halo-retrieval} are {\bf less significant}: Compared to Tab.~\ref{tab1:M_pcorr} the Bayesian learning improves only to $M=1328\rightarrow 1385$ by less than 5 percent,
  whereas BCPNN-WTA does not improve at all ($M=1433\rightarrow 1423$).
\end{itemize}  
%
\begin{table}[ht]
\makebox[\linewidth]{\footnotesize
\begin{tabular}{ccc||cccc}
Core-retrieval, 5step  \\
$\alpha$   &$\alpha k$& $\beta$ & B-WTA             & BCPNN-WTA        & BCPNN2-WTA       & BCPNN3-WTA       \\ \hline
0.78125    &   25     &  0.01   & 1347              &  1418            &  1368            &   1367           \\
0.8125     &   26     &  0.01   & 1345              &  1427            &  1374            &   1372           \\
0.84375    &   27     &  0.01   & 1394              &  1442            &  1409            &   1409           \\
0.875      &   28     &  0.01   & 1433              &  1452            &  1435            &   1434           \\
0.90625    &   29     &  0.01   & 1502              &  1487            &  1488            &   1489           \\
0.9375     &   30     &  0.01   & 1580              &  1507            &  1534            &   1536           \\
           &          &  0.1    & 1575              &  1461            &  1525            &   1530   \\
           &          &  0.001  & 1577              &  1512            &  1532            &   1534   \\
           &          &  0.0001 & 1577              &  1511            &  1530            &   1530   \\
           &          &  0.00001& 1564              &  1505            &  1521            &   1522   \\
           &          &  0      & 1574              &  1467            &  1523            &   1529   \\
0.96875    &   31     &  0.01   & 1589              &  1502            &  1555            &   1554           \\
           &          &  0.1    & 1589              &  1464            &  1548            &   1549   \\
           &          &  0.001  & \underline{1593}  &  \underline{1517}&  \underline{1559}&   \underline{1555}   \\
           &          &  0.0001 & 1592              &  1510            &  1551            &   \underline{1555}  \\
           &          &  0.00001& 1573              &  1510            &  1542            &   1542       \\
           &          &  0      & 1579              &  1425            &  1522            &   1546   \\
\hline
Halo-retrieval, 5step \\
$\alpha$   &$\alpha k$& $\beta$ & B-WTA             & BCPNN-WTA        & BCPNN2-WTA       & BCPNN3-WTA       \\ \hline
1.0312     &   33     &  0.01   & 1385              &  1423            &  1403      &   1408           \\
1.0625     &   34     &  0.01   & 1322              &  1366            &  1330      &   1334           \\
1.0938     &   35     &  0.01   & 1243              &  1339            &  1271      &   1273           \\
1.125      &   36     &  0.01   & 1186              &  1329            &  1230      &   1232           \\
1.1562     &   37     &  0.01   & 1138              &  1320            &  1210      &   1208           \\
1.1875     &   38     &  0.01   & 1104              &  1314            &  1195      &   1192           \\
1.2188     &   39     &  0.01   &$<$1100            &  1315            &  1177      &   1176           \\ 
\hline
\end{tabular}}
\caption{\label{tab5:ANE_maxM_corehalo}
  Estimations of {\bf maximal pattern capacity} $M$ at $p_{\mathrm{corr}}=0.9$ for {\bf ANE} with $K$-WTA {\bf core-retrieval} ((\ref{eq:coreretrieval}); upper part; $\alpha<1$) and {\bf halo-retrieval} ((\ref{eq:haloretrieval}); lower part; $\alpha>1$).
  Parameter $\alpha$ (first column) determines number $\alpha k$ of active units in retrieval step 1 (second column). Parameter $\beta$ corresponds to noise estimation in the later retrieval steps $3,4,\ldots$.
  Networks stored {\bf Palm patterns} ($n=1024,k=32$ with input noise $\lambda=0.9$, $\kappa=0.1$). Note that here the {\bf Bayesian rule reaches the maximal capacity} (B-WTA, $M=1593$) over all experiments so far.
}
\end{table}

We may want to get a closer look on how {\bf output noise evolves over iteration steps}, and whether pattern capacity further increases for more iteration steps: Table~\ref{tab6:ANE_maxM_corehalo_steps}
the results of corresponding experiments where pattern capacity $M_{0.9}$ and mean component faults $f_{10}/f_{01}$ have been evaluated in each iteration step $t\in\{1,2,\ldots,7,10,100\}$
of core-retrieval ($\alpha\in\{0.96875,0.9375\},\beta=0.001$).
As expected from the design of core-retrieval (\ref{eq:coreretrieval}), the output noise in the first step $t=1$ is asymmetric, where number of false-negatives $f_{10}\approx 1-\alpha$ is much higher than
false-positives $f_{01}\approx 0$. {\bf Surprisingly}, in the second step, output noise is still a bit asymmetric with false-positive domination $f_{01}>f_{10}$ (in spite of $K$-WTA with $\alpha=1$, $K=k$).
The reason is that neurons often have identical dendritic potentials equal to the firing threshold and, without symmetry breaking, all these neurons get activated.
\footnote{
  Our implementation of $K$-WTA sorts the neurons' dendritic potentials (in ascending order) and then sets the firing threshold to the $K$th neurons dendritic potential.
  If neuron $K+1$, $K+2$ have the same dendritic potential as neuron $K$, then these neurons will fire as well.
  If we think of $K$-WTA realized by some recurrent inhibition, a similar effect is expected to occur as well in brain networks.
}
This occurs in particular in the {\bf 2nd step} of core-retrieval (\ref{eq:coreretrieval}),
because the estimates $\lambda^{\mathrm{est}}(2):=\alpha=1-p_{10}\approx 1$, $\kappa^{\mathrm{est}}(2):=0=p_{01}$ will cause many very strongly negative synaptic weights $w_{ij}\rightarrow -\infty$
for all learning rules (\ref{eq:wij_Bayesian},\ref{eq:wij_noisy_BCPNN},\ref{eq:wij_Bayesian_BCPNNII_noisy},\ref{eq:wijBCPNNIII}). This inhibition-dominated regime
is similar as described previously for core-retrieval (or ``pattern-part-retrieval'' with $\kappa=0$) in the Willshaw model \cite{Knoblauch:2007_b,Knoblauch_IAMCOIL:HRI2008,Knoblauch_etal_cosyne:2008,Knoblauch/Palm/Sommer:NeurComp2010,Knoblauch:NeurComp2011}: On the one-hand the inhibitory regime is very efficient by Bayesian exclusion
of output neurons becoming active from single strongly inhibitory inputs, but, on the other side, it is prone to $K$-WTA's inability of symmetry-breaking in such a regime.
\footnote{
  In the inhibition-dominated regime, the finite synapses become negligible compared to the (close-to) $-\infty$ weights. Therefore, activation of the neurons depends only on the number
  of infinite synapses a neuron receives, which is typically a small integer (e.g., 0-5). Thus, many neurons will get the same number of $-\infty$ inputs, and therefore a $K$-WTA selection
  will often activate more than $K$ neurons.
}
From the {\bf 3rd step} on, output noise is again symmetric, as expected, and evolves quickly towards its minim, corresponding to maximal pattern capacity.
{\bf Maximal pattern capacity} is typically achieved in steps 5-10. The {\bf Bayesian learning rule} achieves the {\bf maximal pattern capacity} $M=1603$ in step $t=6$, slightly above the previous
maximum at step 5 (cf., Tab.~\ref{tab5:ANE_maxM_corehalo}).
Output noise seems to oscillate, where (for $\alpha=0.96875$) retrieval quality may be a bit better for even $t$ compared to odd $t$, but I have not tested the significance of this effect.
\footnote{
   Such an oscillation may be induced by the asymmetries of output noise in steps 1 and 2. 
}

%
\begin{table}[ht]
\makebox[\linewidth]{\footnotesize
\begin{tabular}{c|cc|cc|cc|cc}
Core-retrieval, \\
$\alpha=0.96875$, \\
$\beta=0.001$  &            &                & BCPNN           &                 & BCPNN2        &               & BCPNN3  \\
step $t$   &B-WTA           &                & -WTA             &                 & -WTA            &               & -WTA        \\ \hline
           & $M_{0.9}$       & $f_{10}/f_{01}$  &   $M_{0.9}$      & $f_{10}/f_{01}$   &  $M_{0.9}$       & $f_{10}/f_{01}$ & $M_{0.9}$       & $f_{10}/f_{01}$  \\ \hline
1          &    0           & 1.0845/0.0845  &    0           &  1.2042/0.2042   &   0            & 1.0935/0.0935 &    0           &   1.0919/0.0919 \\
2          &  1571          & 0.1141/0.1749  &  1411          &  0.2223/0.9347   & 1514           & 0.1386/0.5050 &  1538          &   0.1398/0.1726 \\
3          &  1576          & 0.1204/0.1204  &  1456          &  0.2674/0.2674   & 1541           & 0.1595/0.1595 &  1544          &   0.1412/0.1412 \\
4          &  1597          & 0.1081/0.1081  &  1512          &  0.1851/0.1851   & 1560           & 0.1309/0.1309 &  1556          &   0.1289/0.1289 \\
5          &  1593          & 0.1112/0.1112  &  1517          &  0.1862/0.1862   & 1559           & 0.1387/0.1387 &  1555          &   0.1334/0.1334 \\
6          &\underline{1603}& 0.1058/0.1058  &  1509          &  0.1811/0.1811   & 1557           & 0.1350/0.1350 &  1553          &   0.1303/0.1303 \\
7          &  1591          & 0.1113/0.1113  &\underline{1518}&  0.1858/0.1858   & 1557           & 0.1401/0.1401 &  1552          &   0.1351/0.1351 \\
10         &  1601          & 0.1062/0.1062  &  1516          &  0.1743/0.1743   &\underline{1561}& 0.1327/0.1327 &\underline{1557}&   0.1289/0.1289 \\
100        &  1603          & 0.1058/0.1058  &  1509          &  0.1811/0.1811   & 1557           & 0.1350/0.1350 &  1553          &   0.1303/0.1303 \\
\hline
$\alpha=0.9375$, \\
$\beta=0.001$\\
1          &    0           & 2.0234/0.0234  &    0           &  2.0980/0.0980   &    0           & 2.0340/0.0340 &    0           &   2.0330/0.0330 \\
2          &  1572          & 0.1142/0.1322  &  1456          &  0.2105/0.4965   &  1519          & 0.1542/0.2688 &  1526          &   0.1528/0.1694 \\
3          &  1568          & 0.1208/0.1208  &  1472          &  0.2261/0.2261   &  1519          & 0.1593/0.1593 &  1523          &   0.1487/0.1487 \\
4          &  1572          & 0.1203/0.1203  &  1510          &  0.1858/0.1858   &  1530          & 0.1553/0.1553 &  1530          &   0.1505/0.1505 \\
5          &  1577          & 0.1149/0.1149  &  1512          &  0.1805/0.1805   &  1532          & 0.1526/0.1526 &  1534          &   0.1473/0.1473 \\
6          &  1579          & 0.1144/0.1144  &  1507          &  0.1764/0.1764   &  1527          & 0.1481/0.1481 &  1526          &   0.1480/0.1480 \\
7          &  1569          & 0.1244/0.1244  &  1514          &  0.1829/0.1829   &  1533          & 0.1557/0.1557 &  1532          &   0.1531/0.1531 \\
10         &  1577          & 0.1167/0.1167  &  1513          &  0.1756/0.1756   &  1529          & 0.1501/0.1501 &  1532          &   0.1457/0.1457 \\
100        &  1570          & 0.1191/0.1191  &  1512          &  0.1777/0.1777   &  1532          & 0.1523/0.1523 &  1532          &   0.1501/0.1501 \\
\hline
\end{tabular}}
\caption{\label{tab6:ANE_maxM_corehalo_steps}
  Estimations of {\bf maximal pattern capacity} $M_{0.9}$ at $p_{\mathrm{corr}}=0.9$
  (interpolated between samplings at $M=1100,1200,\ldots,1900$)
  and mean false-negative and false-positive {\bf component errors} $f_{10}$ and $f_{01}$
  (for fixed $M=1600$, close to the maximum of the Bayesian learning rule)
  as functions of {\bf iteration steps}
  for {\bf ANE} with $K$-WTA {\bf core-retrieval} (\ref{eq:coreretrieval}) of {\bf Palm patterns} using $\beta=0.001$ with $\alpha=0.9375$ (upper part)
  and $\alpha=0.96875$ (lower part), corresponding to the {\bf two best cases} of Tab.~\ref{tab5:ANE_maxM_corehalo}.
}
\end{table}

\subsection{ANE with core- and halo-retrieval for Willshaw patterns}\label{sec:experiments_results_ANE_corehalo_Willshaw}
Core- and halo-retrieval can also be realized for Willshaw patterns. Interpreting dendritic potentials $x_j:=\ln(\frak{r}_j)=\ln\frac{\pr[u_j=1]}{\pr[u_j=0]}$ as log-odds-ratios
as in (\ref{eq:xj_Bayesian}) with (\ref{eq:fpp}), we may redefine $\alpha$ for Willshaw patterns
as the {\bf inverse decision threshold on the odds-ratio},
\begin{align}
  \hat{u}_j=\begin{cases}
               1 &, \frak{r}_j\ge\frac{1}{\alpha}\\
               0 &, \mathrm{else}
            \end{cases}    =
            \begin{cases}
               1 &, x_j:=\ln\frak{r}_j \ge \theta_j:=\ln\frac{1}{\alpha}=-\ln\alpha \\
               0 &, \mathrm{else}
            \end{cases}
  \label{eq:inverse_alpha_decision_threshold}
\end{align}
where $\theta_j=-\ln\alpha$ is the firing threshold of the neuron. Then $\alpha$ has a similar role for Willshaw patterns (with fixed firing thresholds) as for K-WTA on Palm patterns:
For the default case $\alpha=1$ the firing threshold is $\theta_j=0$
aiming to activate on average all $k$ neurons of a pattern. For $\alpha>1$ we have a lower firing threshold $\theta_j<0$ such that,  on average, more than $k$ neurons get activated, corresponding to halo-retrieval.
And for $\alpha<1$ we have $\theta_j>0$ and on average less than $k$ neurons get activated, corresponding to core-retrieval. However, predicting from $\alpha$ precisely how many neurons will get activated is
more difficult.
\footnote{
  In principle, we could estimate the number of active neurons by making a Gaussian assumption on the dendritic potentials $x_j$ \cite{Knoblauch:NeurComp2011}. However, the Gaussian assumption
  becomes strictly valid only in the limit $Mp^2\rightarrow\infty$, whereas for finite networks ($M\ll\infty$) we cannot expect precise results
  (cf., imprecise theoretical estimates based on Gaussian assumption in Figs.~\ref{fig1:ExpRepro2A}-\ref{fig3:CompLan_M001}).
}
Similar to the numerical experiments of Tabs.~\ref{tab5:ANE_maxM_corehalo},\ref{tab6:ANE_maxM_corehalo_steps}, we still can test different values of $\alpha(t)$
and estimate the corresponding mean false negative $f_{10}(t)$ and false positive $f_{01}(t)$ numbers for relevant retrieval steps $t$. In analogy to (\ref{eq:coreretrieval},\ref{eq:haloretrieval})
this leads to the following {\bf core-retrieval} schedule for Willshaw patterns,
\begin{align}
  &\mbox{Iteration step 1:}           &\lambda^{\mathrm{est}}&:=\lambda, \quad &\kappa^{\mathrm{est}}&:=\kappa, \quad &\alpha&<1 \nonumber\\
  &\mbox{Iteration step 2:}           &\lambda^{\mathrm{est}}&:=1-\frac{f_{10}(1)}{k} , \quad &\kappa^{\mathrm{est}}&:=0,      \quad &\alpha&:=1 \nonumber\\
  &\mbox{Iteration steps 3,4,\ldots:} &\lambda^{\mathrm{est}}&:=1-\beta, \quad &\kappa^{\mathrm{est}}&:=\beta,  \quad &\alpha&:=1 \label{eq:coreretrieval_Willshaw}
\end{align}
and the following {\bf halo-retrieval} schedule for Willshaw patterns
\begin{align}
  &\mbox{Iteration step 1:}           &\lambda^{\mathrm{est}}&:=\lambda, \quad &\kappa^{\mathrm{est}}&:=\kappa, \quad &\alpha&>1 \nonumber\\
  &\mbox{Iteration step 2:}           &\lambda^{\mathrm{est}}&:=0 , \quad &\kappa^{\mathrm{est}}&:=\frac{f_{01}}{k},      \quad &\alpha&:=1 \nonumber\\
  &\mbox{Iteration steps 3,4,\ldots:} &\lambda^{\mathrm{est}}&:=1-\beta, \quad &\kappa^{\mathrm{est}}&:=\beta,  \quad &\alpha&:=1 \label{eq:haloretrieval_Willshaw}
\end{align}
Tab.~\ref{tab7:ANE_maxM_Willshaw_core} shows some results from {\bf experiments with core-retrieval for Willshaw patterns}:
Instead of estimating $f_{10}(1)$ in each experiment, as formally required in (\ref{eq:coreretrieval_Willshaw}), for simplicity,
we optimized $\lambda^{\mathrm{est}}(2)$ in the second retrieval step just as a regular hyperparameter (first column), in conjunction with $\alpha(1)$ in the first retrieval step.
For a wide range of $\lambda^{\mathrm{est}}(2)\in[0.7;0.99]$ optimal $\alpha(1)$ was close to 0.3 (shown only for $\lambda^{\mathrm{est}}(2)\in\{0.97,0.85\}$), and optimal
$\beta$ close to $0.01$ (shown only for $\lambda^{\mathrm{est}}(2)=0.97$). The overall {\bf maximum capacity} was $M=1222$ for the {\bf Bayesian learning rule}, significantly (almost 10 percent) above
the previous maximum for Willshaw patterns for fixed low noise estimations ($M=1115$; see Tab.~\ref{tab1:M_pcorr}). By contrast,
for BCPNN I did not observe any improvements (max. $M=1029$ vs. $M=1095$ in Tab.~\ref{tab1:M_pcorr}). 
As the maximum for BCPNN seems to require smaller $\alpha$ and $\beta$ than for the Bayesian rule, one may suspect that it coincides with the fixed low noise limit of Tab.~\ref{tab1:M_pcorr}. 

%
\begin{table}[ht]
\makebox[\linewidth]{\footnotesize
\begin{tabular}{cccc||cccc}
Core-retrieval  \\
$\lambda^{\mathrm{est}}(2)$&$\alpha$(1) & $\beta$  & steps& B                 & BCPNN            & BCPNN2           & BCPNN3           \\ \hline
 0.99                  & 0.3        & 0.01    &   5  & 1153              & 969             & 1077             & 1033 \\
 0.98                  &            &         &      & 1167              & 972             & 1100             & 1039 \\
 0.97                  & 0.01       &         &      & 1087              & \underline{1029}& 1019             & 1047 \\
                       & 0.1        &         &      & 1164              & 1013            & 1100             & 1059 \\
                       & 0.2        &         &      & 1177              & 988             & 1108             & 1038 \\
                       & 0.3        & 0.005   &      & 1184              & \underline{1029}& 1118             & 1068 \\
                       &            & 0.01    &      & 1193              & 963             & 1122             & 1032 \\
                       &            & 0.05    &      & 1158              & $<$900          & 1066             & $<$ 900 \\
                       & 0.4        & 0.01    &      & 1190              & 968             & 1110             & 1030 \\
                       & 0.5        &         &      & 1188              & 950             & 1112             & 1021 \\ 
 0.95                  & 0.3        &         &      & 1205              & 969             & 1128             & 1031 \\
 0.93                  &            &         &      & 1216              & 978             & 1146             & 1038 \\
 0.9                   &            &         &      & 1212              & 970             & 1149             & 1034 \\
 0.85                  & 0.1        &         &      & 1206              & 1012            & 1154             & 1055 \\
                       & 0.2        &         &      & 1214              & 985             & 1157             & 1043 \\
                       & 0.3        &         &  1   & $<$900            & $<$900          & $<$900           & $<$900 \\
                       &            &         &  2   & 1200              & $<$900          & 1096             & \underline{1072} \\
                       &            &         &  3   & 1217              & $<$900          & 1110             & 1037 \\
                       &            &         &  4   & 1217              & 974             & 1157             & 1039 \\
                       &            &         &  5   & \underline{1222}  & 975             & \underline{1159} & 1035 \\
                       &            &         &  6   & 1218              & 979             & \underline{1159} & 1039 \\
                       &            &         &  7   & 1218              & 978             & \underline{1159} & 1039 \\
                       &            &         &  8   & 1218              & 979             & \underline{1159} & 1039 \\
                       &            &         & 10   & 1218              & 979             & \underline{1159} & 1039 \\
                       & 0.4        &         &  5   & 1221              & 964             & \underline{1159} & 1033 \\
                       & 0.5        &         &      & 1212              & 959             & 1144             & 1020 \\
 0.8                   & 0.3        &         &      & 1216              & 974             & 1156             & 1032 \\
 0.7                   &            &         &      & 1210              & 975             & 1156             & 1038 \\
\end{tabular}}
\caption{\label{tab7:ANE_maxM_Willshaw_core}
  Estimations of {\bf maximal pattern capacity} $M$ at $p_{\mathrm{corr}}=0.9$ for {\bf ANE} with {\bf core-retrieval} (\ref{eq:coreretrieval_Willshaw}) (interpolated between samples at $M=900,1000,\ldots,1500$).
  Networks stored {\bf Willshaw patterns} ($n=1024,k=32$ with input noise $\lambda=0.9$, $\kappa=0.1$).
  Parameter $\lambda^{\mathrm{est}}(2)$ (column 1) estimates pattern completeness before retrieval step 2 (we have not measured $f_{10}(1)$ as defined in (\ref{eq:coreretrieval_Willshaw}),
  but instead just used fixed values for simplicity). Parameter $\alpha$ (column 2) determines active units in step 1 (see text for details).
  Parameter $\beta$ corresponds to noise estimation in the later retrieval steps $3,4,\ldots$. Empty table cells have same values as in the cell above.
}
\end{table}

Similar as for Palm patterns (Tab.~\ref{tab5:ANE_maxM_corehalo}), {\bf halo retrieval for Willshaw patterns did not improve results} compared to Tab.~\ref{tab1:M_pcorr} or compared to core-retrieval (Tab.~\ref{tab7:ANE_maxM_Willshaw_core}).
In a preliminary explorative optimization, I tested  $\kappa^{\mathrm{est}}(2)\in\{0.05, 0.1, 0.15, 0.2, 0.3\}$, $\alpha(1)\in\{1.1,1.2,1.5,1.6,\ldots,1.9\}$ for $\beta=0.01$ for 5 retrieval steps (data not shown).
The best result for Bayesian learning was $M=1019$ at $p_{\mathrm{corr}}=0.9$ for smallest values $\kappa^{\mathrm{est}}(2)=0.05$ and $\alpha(1)=1.1$ (and the other learning rule had always even lower capacities).
As this is significantly below the optimum for fixed noise estimates ($M=1115$; see Tab.~\ref{tab1:M_pcorr}), I did not further explore halo retrieval here.
\footnote{
  Indeed, capacities for halo retrieval with Willshaw patterns seem to increase for $\kappa^{\mathrm{est}}(2)\rightarrow 0$ and $\alpha(1)\rightarrow 1$ which corresponds
  to the fixed low noise estimates limit of Tab.~\ref{tab1:M_pcorr}.
}

\section{Summary and discussion}\label{sec:discussion}
This technical report detailed on the auto-associative variant of the ``optimal'' Bayesian learning rule (\ref{eq:wij_Bayesian},\ref{eq:xj_Bayesian})
for neural associative networks \cite{Knoblauch:NeurComp2011}
and compared it to several variants of the BCPNN learning rule,
\begin{itemize}
\item original BCPNN (\ref{eq:wij_Bayesian_BCPNN_weights},\ref{eq:xj_Bayesian_BCPNN_potentials}) \cite{Lansner/Ekeberg:1987,Lansner/Ekeberg:1989}:
      without noise estimation (implicitly assuming $p_{01}=p_{10}=0$), without stabilization ($M_{11}'=M_{11}$ or $\eps_s=0$).
\item stabilized BCPNN (\ref{eq:wij_Bayesian_BCPNN_weights_stable}) \cite{Lansner/Holst:1996,Johansson/Lansner:2007,Martinez:BCPNN:2022,Lansner/Ravichandran/Herman:2023}:
      with numerical stabilization to prevent infinite synaptic weights. 
\item noisy BCPNN (\ref{eq:wij_noisy_BCPNN}) \cite{Knoblauch:NeurComp2011}: with estimation of input noise, without stabilization.
\item noisy BCPNN2 (\ref{eq:wij_Bayesian_BCPNNII_noisy}) \cite{Knoblauch:NeurComp2011}: with estimation of input noise, like BCPNN, but considers both 1 and 0 components.
\item noisy BCPNN3 (\ref{eq:wijBCPNNIII}) \cite{Knoblauch:NeurComp2011}: with estimation of input noise, like BCPNN, but based on odds-ratio.
\end{itemize}
and additionally to noisy BCPNN2 (\ref{eq:wij_Bayesian_BCPNNII_noisy}) and noisy BCPNN3 (\ref{eq:wijBCPNNIII}) from \cite{Knoblauch:NeurComp2011}.
By replacing $M_{11}$ by $M_{11}'$ from (\ref{eq:wij_Bayesian_BCPNN_weights_stable}),
the stabilization method has also be tested
with the Bayesian rule and BCPNN2/3. From the numerical experiments with these rules we can draw the following major conclusions:
\begin{enumerate}
\item As one major purpose of this study, we can {\bf confirm the correctness} of the implementation of the {\bf Bayesian learning rule} of \cite{Knoblauch:NeurComp2011}.
  The reported {\bf anomalies} that BCPNN can perform sometimes better than the ``optimal'' Bayesian rule \cite{Lansner:2024:perscomm} can be explained in terms of
  {\bf violations} of the {\bf pre-assumptions} of the Bayesian learning rule (independent random pattern components, fixed identical noise probabilities $p_{01},p_{10}$ for all neurons
  and all retrieval steps).
\item In contrast to one-step retrieval, the {\bf Bayesian learning rule cannot be guaranteed to be optimal} during {\bf iterative retrieval}:
  As iterative retrieval reduces noise
  over iteration steps, there cannot be fixed noise estimates that are valid over all iteration steps. If noise estimates $p_{01},p_{10}$ or
  $\lambda^{\mathrm{est}}, \kappa^{\mathrm{est}}$ must be fixed, then it is often beneficial to use smaller values than the initial input noise that match approximately
  final output noise in the retrieval result (Tab.~\ref{tab1:M_pcorr}).
\item {\bf Stabilized BCPNN} or {\bf BCPNN with low noise estimates} (which has a similar effect as stabilization) seem indeed to be {\bf superior over the Bayesian rule}
  for {\bf iterative retrieval} with {\bf Palm patterns} (constant number $k$ of active units per pattern) and $k$-WTA. The reason seems to be that
  BCPNN seems to better able to generalize to the various noise levels during iterative retrieval.
\item By contrast, for {\bf Willshaw patterns} (i.i.d. pattern components, where active units per pattern is binomially distributed),
  the {\bf Bayesian learning was generally superior over BCPNN}. The reason is that $K$-WTA cannot be applied for Willshaw patterns, and
  BCPNN cannot easily predict the correct firing thresholds necessary for optimal retrieval results.
\item The idea of {\bf Adaptive Noise Estimation (ANE)} is to adapt input noise estimates in each step over the course of iterative retrieval,
  however, leading only to moderate improvements for the Bayesian learning rule. The reason is that mean noise estimates per step still does not account
  for large variations per network, pattern and components.
\item {\bf ANE with core or halo retrieval} tries to solve this problem by ``designing'' retrieval conditions
  where pattern noise over iterative retrieval becomes predictable with low variations. In particular, {\bf core retrieval} can strongly increase storage capacity
  for the Bayesian learning rule by applying high firing thresholds in the first retrieval step, such that only false negative component errors occur.
  With this mechanism, the {\bf Bayesian learning rule significantly exceeds capacity of all BCPNN versions}, both for Palm and Willshaw patterns. 
\end{enumerate}
Tab.~\ref{tab8:ComparisonSummary} gives a {\bf summary of maximal pattern capacities} (at 90 percent correct) for $n=1024, k=32$ for Bayesian and BCPNN learning rules under different conditions (Palm/Willshaw patterns and ZNE,FNE,ANE, see caption). The {\bf overall maximum capacity} $M=1603$ is achieved by the Bayesian rule for {\bf Palm patterns} (fixed number of active units per pattern) and ANE (adaptive noise estimation),
requiring synapses that can change weights over iterative retrieval steps. For biologically more plausible conditions like fixed synaptic weights (e.g., for zero noise estimates, ZNE)
the BCPNN rule achieves the best pattern capacity $M=1439$ for Palm patterns. For {\bf Willshaw patterns} (variable number of active units per pattern) the Bayesian rule is again superior, at least for
synapses including fixed or adaptive noise estimates ($M=1115$ or $M=1222$), where for the most conservative condition ZNE (zero noise estimates), the BCPNN rule still performs reasonably well ($M=1102$).
In summary, the Bayesian rule yields maximum pattern capacity under specialized conditions (possibility of adaptive noise estimation), whereas (stabilized) BCPNN seems more robust under simpler conditions
without noise estimation. 

\begin{table}[ht]
\makebox[\linewidth]{\footnotesize
\begin{tabular}{c||ccc|ccc}
            & Bayes                                                  &                                     &                                                     & BCPNN  &       &      \\ 
            & ZNE                                                    & FNE                                 &  ANE                                                & ZNE                                 & FNE & ANE \\ \hline\hline
  Palm/K-WTA      & 0                                                & 1335                                & \underline{1603}                                    & 1439                                 & 1439    & 1518\\    
                  &(Tabs.~\ref{tab1:M_pcorr},\ref{tab2:M_pcorr_stab})&(stab., Tab.~\ref{tab2:M_pcorr_stab})&(core-retr., Tab.~\ref{tab6:ANE_maxM_corehalo_steps})& (stab., Tab.~\ref{tab2:M_pcorr_stab})&(see ZNE)&(core-retr., Tab.~\ref{tab6:ANE_maxM_corehalo_steps}) \\ \hline
Willshaw/threshold& 0                                               & 1115                                 & \underline{1222}                                    & 1102                                & 1102    & 1102 \\ 
                  &(Tabs.~\ref{tab1:M_pcorr},\ref{tab2:M_pcorr_stab} &(Tab.~\ref{tab1:M_pcorr})            &(core-retr., Tab.~\ref{tab7:ANE_maxM_Willshaw_core}) &(stab., Tab.~\ref{tab2:M_pcorr_stab})&(see ZNE)&(see ZNA) \\
\hline
\end{tabular}}
\caption{\label{tab8:ComparisonSummary}
  {\bf Summary/comparison} of experimentally estimated {\bf maximal pattern capacity} $M_{0.9}$ at correctness level $p_{\mathrm{corr}}=0.9$
  for network size $n=1024$ with (average) $k=32$ active units (and 20 percent input noise $\lambda=0.9$, $\kappa=0.1$) for Bayesian and BCPNN learning rules under different conditions.
  {\bf Palm/K-WTA} means that all patterns are random with exactly $k$ active units (where noisy query input patterns have $\mathrm{round}(\lambda k)=29$ correct and $\mathrm{round}(\kappa k)=3$ false ones)
  and retrieval output patterns are determined by K-winners-take-all decisions.
  {\bf Willshaw/threshold} means that all pattern have i.i.d. components with average $k$ active units (and noisy query input patterns have on average $\lambda k=28.8$ correct and $\kappa k=3.2$ false ones)
  and retrieval output patterns are determined by a fixed firing threshold.
  {\bf ZNE} means ``zero-noise-estimates'' setting noise estimates $p_{01}=p_{10}=0$ for all synapses (and biases) to zero.
  {\bf FNE} means ``fixed-noise-estimates'' where $p_{01}$ and $p_{10}$ are identical constant for all synapses (and biases). 
  {\bf ANE} means ``adaptive-noise-estimates'' where $p_{01}(t)$ and $p_{10}(t)$ are identical for all synapses (and biases), but can vary over retrieval steps $t=1,2,\ldots$. 
}
\end{table}

As ANE can significantly increase storage capacity, we may want to to further explore both the usability for technical applications and the relevance for neurophysiological models:
Concerning {\bf technical applications}, a {\bf naive implementation of ANE} would require considerably more memory for synaptic weights, that is, one for each set of noise estimates. For example,
implementing the most efficient core-retrieval-schedule (\ref{eq:coreretrieval}) would require three weight matrixes for the noise estimates in step 1, step 2, and the further steps 3,4,\ldots.
By contrast, a {\bf memory-efficient implementation of ANE} may store only the synaptic counters $M_1$ and $M_{11}$ (see end of section~\ref{sec:learning}) and compute synaptic weights
like (\ref{eq:wij_Bayesian}) ``on the fly'' in each retrieval step.
\footnote{
  It may be beneficial to use look-up-tables with pre-computed values of the log-function, as evaluations are $\log(.)$ are computationally expensive. 
}
For sparse activity, only a small number of synapses must be re-computed in each retrieval step. The bias (\ref{eq:xj_Bayesian})
could still be pre-computed and stored three times, which is still negligible compared to the memory requirements for the weight matrix ($O(n)$ vs. $O(n^2)$). While I have studied here only the most 
random pattern and network types, it may be possible to increase storage capacity for specialized pattern structures and retrieval strategies.
For example, previous studies on the Willshaw/Steinbuch network-type \cite{Steinbuch:1961,Willshaw/Buneman/Longuet-Higgins:1969,Palm:1980}
have shown the benefits of AND-ing between ``halo''-retrieval steps (see LK+ strategy in \cite{Schwenker/Sommer/Palm:1996})
or introduced even more advanced retrieval strategies for ``hypercolumnar'' block patterns \cite{Gripon/Berrou:2011,Gripon/Berrou:2012,Yao/Gripon/Rabbat:2014,Gripon/Heusel/Lowe/Vermet:2016,KnoblauchPalm:NeurComp2020}. Future works should investigate
how these strategies could be adapted to increase storage capacity of Bayesian or BCPNN-type associative memory.

Concerning {\bf neurobiological plausibility of ANE}, we may try to relate ANE to synaptic modulation mechanism that can change synaptic efficacy on a short time scale, e.g., of tens to hundreds of milliseconds
\footnote{
  A common {\bf neurophysiological interpretation} of iterative retrieval in associative networks is that they may correspond to phases of enhanced gamma-band oscillations (25-60 Hz) observed in the brain related
  to cognitive events (like memory retrieval, attention, or visual recognition of objects \cite{Gray/Konig/Engel/Singer:1989,Gray:1999,Fries/Reynolds/Rorie/Desimone:2001})
  and lasting for several hundred milliseconds \cite{Knoblauch/Palm:2001_a,Knoblauch/Palm:2002_a,Knoblauch/Palm:2002_b}. Here each gamma cycle may correspond to an individual retrieval step, whereas
  iterative retrieval may correspond to the phases of multiple (e.g., 5-15) gamma cycles. As a consequence, a possible ANE mechanism should also work on this same time scale of tens to hundreds of milliseconds.
}
One candidate mechanism would be {\bf short-term-plasticity (STP)} like synaptic facilitation and depression \cite{Zucker/Regehr:2002,Hennig:2013}. This is usually thought of as an activity dependent modulation process, where
synaptic weights can either increase or decrease following presynaptic activity. However, ANE would rather require a ``global'' mechanism that modulates synaptic efficiency of all synapses,
independent of previous neural activity. It may still be possible that STP could approximate ANE as, over the course of iterative retrieval mostly the same neurons are active, and STP would act equivalently
to a global modulation of all synapses of the network. Future work should investigate the feasibility of such STP-based models of ANE more thoroughly.

\subsubsection*{Acknowledgments} I am grateful to Anders Lansner for stimulating this research by valuable discussions on BCPNN and optimal Bayesian learning.

%
%

%
%
\setlength{\itemsep}{0cm}
\bibliographystyle{plain}
\bibliography{../../../../03-resources/LATEX_gen/neuroAK,../literatur}

\end{document}